\documentclass[pdflatex, sn-basic, iicol]{sn-jnl}%

\usepackage{graphicx}%

\usepackage{mathtools}
\usepackage{multirow}%
\usepackage{amsmath,amssymb,amsfonts}%
\usepackage{amsthm}%
\usepackage{adjustbox}%
\usepackage{mathrsfs}%
\usepackage{comment}%
\usepackage{tabularx}
\usepackage[title]{appendix}%
\usepackage{xcolor}%
\usepackage{array}
\usepackage{textcomp}%
\usepackage{manyfoot}%
\usepackage{booktabs}%
\usepackage{algorithm}%
\usepackage{colortbl}
\usepackage{algorithmicx}%
\usepackage{algpseudocode}%
\usepackage{listings}%
\usepackage{svg}%
\usepackage{caption}
\usepackage{float}
\usepackage{lipsum}
\usepackage{amsmath}
\usepackage{import}%
\usepackage{adjustbox}%
\usepackage{xr}
\usepackage[accsupp]{axessibility} %
\usepackage{dsfont}

\theoremstyle{thmstyleone}%

\theoremstyle{thmstyletwo}%

\theoremstyle{thmstylethree}%

\definecolor{bestblue}{HTML}{3a7bc1}
\definecolor{bestgreen}{HTML}{41914f}
\newcommand{\loss}{\mathcal{L}}

\begin{document}

\title[Article Title]{\texorpdfstring{\begin{tabular}{c}Multi-view Surface Reconstruction \\ Using Normal and Reflectance Cues\end{tabular}}{Multi-view Surface Reconstruction Using Normal and Reflectance Cues}}

\author*[1]{\fnm{Robin} \sur{Bruneau}}\email{robin.bruneau@uzh.ch}
\equalcont{These authors contributed equally to this work.}

\author*[2]{\fnm{Baptiste} \sur{Brument}}\email{baptiste.brument@irit.fr}
\equalcont{These authors contributed equally to this work.}

\author[3]{\fnm{Yvain} \sur{Quéau}}\email{yvain.queau@ensicaen.fr}

\author[2,4]{\fnm{Jean} \sur{Mélou}}\email{jean.melou@irit.fr}

\author[5]{\fnm{François} \sur{Bernard Lauze}}\email{francois@di.ku.dk}

\author[2]{\fnm{Jean-Denis} \sur{Durou}}\email{jean-denis.durou@irit.fr}

\author[6]{\fnm{Lilian} \sur{Calvet}}\email{lilian.calvet@balgrist.ch}

\affil[1]{\orgdiv{DQBM}, \orgname{University of Zurich}, \orgaddress{\country{Switzerland}}}

\affil[2]{\orgdiv{IRIT, UMR CNRS 5505}, \orgname{Université de Toulouse}, \orgaddress{\country{France}}}

\affil[3]{\orgdiv{GREYC}, \orgname{CNRS, UNICAEN, ENSICAEN, Normandie Universit\'e}, \orgaddress{\country{France}}}

\affil[4]{\orgdiv{FittingBox}, \orgaddress{Toulouse, \country{France}}}

\affil[5]{\orgdiv{DIKU}, \orgname{University of Copenhagen}, \orgaddress{\country{Denmark}}}

\affil[6]{\orgdiv{ROCS, University of Zurich}, \orgname{OR-X, Balgrist University Hospital}, \orgaddress{\country{Switzerland}}}

\abstract{
Achieving high-fidelity 3D surface reconstruction while preserving fine details remains challenging, especially in the presence of materials  with complex reflectance properties and without a dense-view setup. In this paper, we introduce a versatile framework that incorporates multi-view normal and optionally reflectance maps into radiance-based surface reconstruction. Our approach employs a pixel-wise joint re-parametrisation of reflectance and surface normals, representing them as a vector of radiances under simulated, varying illumination. This formulation enables seamless incorporation into standard surface reconstruction pipelines, such as traditional multi-view stereo (MVS) frameworks or modern neural volume rendering (NVR) ones. Combined with the latter, our approach achieves state-of-the-art performance on multi-view 
photometric stereo (MVPS) benchmark datasets, including DiLiGenT-MV, LUCES-MV and Skoltech3D. In particular, our method excels in reconstructing fine-grained details and handling challenging visibility conditions. The present paper is an extended version of the earlier conference paper by Brument et al (in Proceedings of the IEEE/CVF Conference on Computer Vision and Pattern Recognition (CVPR), 2024), featuring an accelerated and more robust algorithm as well as a broader empirical evaluation. The code and data relative to this article are available at \url{https://github.com/RobinBruneau/RNb-NeuS2}.
}


\maketitle
\clearpage
\section{Introduction}
\label{sec:intro}

Surface reconstruction is essential in various fields, including cultural heritage preservation, medical imaging, virtual and augmented reality, digital twinning, and content creation for games and film production. 
Despite significant advancements, the performance of state-of-the-art surface reconstruction methods remains highly dependent on scene characteristics, notably the presence of fine-scale geometric details and the possibly complex reflectance properties of the surface.

The recovery of fine-grained structures is a long-standing bottleneck in 3D surface reconstruction. Traditional multi-view stereo (MVS) methods \citep{furukawa2007accurate, schonberger2016pixelwise} often produce overly smoothed surfaces, struggling with sharp edges and intricate geometries \citep{seitz2006comparison, FurukawaP15}. 
Recent neural approaches have advanced surface reconstruction in this direction by shifting from local, patch-based optimisation to global, pixel-wise optimisation \citep{yariv2020idr, yariv2021volume, oechsle21unisurf, wang21neus}. Notably, methods such as HF-NeuS~\citep{wang2022hfneus}, PET-NeuS~\citep{wang2023petneus}, NeuS2~\citep{wang23neus2}, and Neuralangelo~\citep{li2023neuralangelo} excel in capturing fine surface details -- at least, when the density of viewpoints is high \citep{brument24rnbneus, logothetis2024luces}. 

On the other hand, approaches leveraging multi-light information have demonstrated strong performance in recovering intricate surface geometries, even in the sparse-view scenario. This is after all the principle of photometric stereo (PS), which achieves monocular 2.5D reconstruction under the form of a normal map, under varying illumination~\citep{woodham1980photometric}. Several recent multi-view, multi-light setups such as SuperNormal~\citep{cao2024supernormal} or our previous work RNb-NeuS~\citep{brument24rnbneus} thus employ PS at each viewpoint, before resorting to multi-view normal integration for complete 3D surface reconstruction. 

Apart from fine details, complex materials (metallic, specular, translucent, rough, etc.) represent another challenge due to the strong view-dependence of reflectance -- hence of radiance, which breaks the brightness consistency assumption upon which MVS is fundamentally based~\citep{seitz2006comparison, FurukawaP15}. 
Although recent single-image approaches have demonstrated promising results in reconstructing accurate surface normals even for complex materials, by leveraging inductive biases and diffusion mechanisms \citep{bae2024dsine, ye2024stablenormal} within vision transformers (ViTs) \citep{dosovitskiy2020vit}, the most natural way to cope with complex materials remains active illumination i.e., PS. 

The most recent deep learning-based PS techniques, also based on ViTs, indeed recover detailed surface normal maps and fine geometric structures even for highly complex materials, both when the illumination is calibrated~\citep{wei2025revisiting} or when it is unknown and possibly spatially-varying (so-called ``universal'' PS setup~\citep{ikehata2022universal, ikehata23sdmunips, hardy2024unimsps}). Besides, some of these methods~\citep{ikehata2022universal, ikehata23sdmunips} even extend beyond normal estimation by estimating reflectance properties such as diffuse colour, roughness, or metalness, thereby enabling virtual relighting applications.

Motivated by the strong performance of recent PS methods in estimating detailed surface normals and reflectance properties, 
the present work proposes a multi-view surface reconstruction approach designed to inherit these advancements. Our approach follows a two-stage decomposition strategy that proves effective 
even without dense multi-view inputs. Specifically, we first estimate per-view surface normals (and optionally reflectance properties) from multi-light input data, and subsequently reconstruct a surface that best aligns with these estimates across multiple views.

\begin{figure*}
    \centering
    \includegraphics[width=\linewidth]{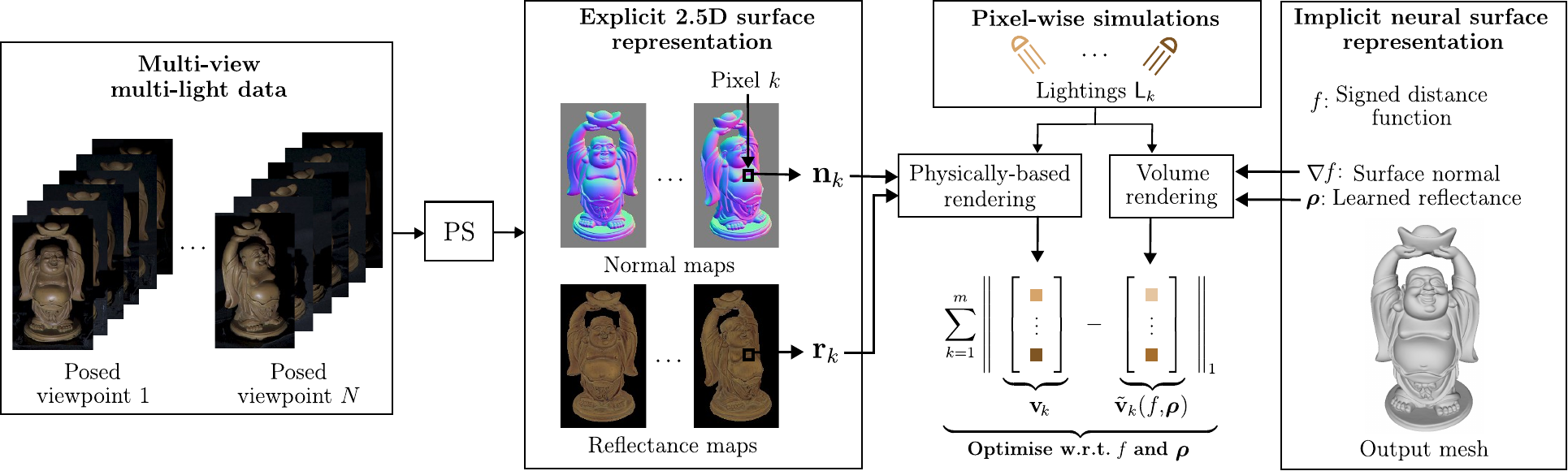}
    \caption{Overview of the surface reconstruction pipeline ``RNb-NeuS'' proposed in~\citep{brument24rnbneus}. From multi-view multi-light data, photometric stereo (noted PS) first estimates per-view normal maps $\{\mathbf{n}_{k}\}$ and optionally reflectance maps $\{\mathbf{r}_{k}\}$. These estimates are re-parametrised into radiance vectors $\mathbf{v}_{k} = \mathsf{F}(\mathbf{n}_{k}, \mathbf{r}_{k}, \mathsf{L}_{k})$ via a physically-based rendering function $\mathsf{F}$ under simulated lightings $\mathsf{L}_{k}$. An implicit neural representation, consisting of a signed distance function $f$ and a reflectance $\boldsymbol{\rho}$, is then learned by minimising the discrepancy between $\mathbf{v}_{k}$ and its volume-rendered counterpart $\tilde{\mathbf{v}}_k(f,\boldsymbol{\rho})$. The final surface mesh is extracted as the zero-level set of $f$ after optimisation.}
    \label{fig:pipeline}
\end{figure*}

This paper builds upon and significantly extends our previous conference publication~\citep{brument24rnbneus}, where the foundational concepts of this two-stage approach were introduced. Therein, 
we presented the key idea to re-parametrise normal and reflectance priors into simulated radiance values on a per-pixel basis, facilitating seamless incorporation into existing surface reconstruction pipelines, notably neural volume rendering (NVR) frameworks based on signed distance functions (SDF). This innovative way to perform multi-view normal (and reflectance) integration yields a state-of-the-art multi-view photometric stereo (MVPS) technique, summarised in Figure~\ref{fig:pipeline}, which we evaluated on the DiLiGenT-MV dataset~\citep{LiZWSDT20} using the SDM-UniPS method~\citep{ikehata23sdmunips} for normal and reflectance estimation.

In the present extension, we substantially improve upon this initial contribution in terms of speed, robustness, versatility, and evaluation. In particular: 
\begin{itemize}
    \item We implement our approach within the NeuS2 framework~\citep{wang23neus2}, achieving a 100$\times$ speed-up compared to our original implementation (see Section~\ref{subsec:rnbneus2_technical}).
    \item We improve the robustness of our approach to reflectance singularities by introducing reflectance embedding (see Section~\ref{subsec:reflectance_singularities}).
    \item In addition to volume rendering, we validate our re-parametrisation in a traditional patch-based MVS framework, demonstrating both its broad applicability and its ability to achieve \textit{exact} integration in a noiseless setup (see Section~\ref{sec:surface_sweeping}). 
    \item An alternative use case is explored, by substituting photometric stereo-estimated normals with ones derived from dense MVS under fixed illumination~\citep{schonberger2016pixelwise}, thereby improving fine-structure reconstruction on the DTU dataset~\citep{jensen2014large} (see Section~\ref{subsec:single_light}).
    \item A much deeper evaluation of our MVPS solution is conducted (see Section~\ref{sec:exp_mvps}), notably under a sparse-view setup and on additional benchmarks such as LUCES-MV~\citep{logothetis2024luces} and Skoltech3D~\citep{voynov2023multi}, using various PS results as input~\citep{ikehata23sdmunips, hardy2024unimsps}. We also compare our results against the very recent SuperNormal method~\citep{cao2024supernormal}, which happens to be a special case of our approach (see Section~\ref{subsec:supernormal_light}).   
\end{itemize}

The rest of this article is organised as follows. We first review the relevant literature in Section~\ref{sec:sota}. Then, we show in Section~\ref{sec:reparametrisation} how to jointly re-parametrise normal and reflectance data into simulated radiance values. We embed this re-parametrisation in a traditional MVS framework in Section~\ref{sec:surface_sweeping}, to demonstrate the feasibility of exact surface reconstruction from normals and reflectance cues. To ensure robustness, in Section~\ref{sec:volume_rendering} we then turn our attention to replacing this framework by a volume rendering one based on neural implicit surfaces. Extensive experiments validating the proposed approach are then conducted in Section~\ref{sec:experimental_results}, before our conclusions are drawn in Section~\ref{sec:conclusion}.

\section{Related Work}
\label{sec:sota}

With our normal and reflectance integration approach in mind, let us review estimation of normals, of reflectance, and integration techniques.   

\subsection{Normal Estimation}

Normal maps carry out high-frequency geometric information. Their estimation is achieved differently depending on the amount of input data.

\textbf{Single-image normal estimation} is highly beneficial for 3D scene reconstruction, since it preserves local geometry without metric ambiguity~\citep{yu2022monosdf, wang2022neuris}. Recent approaches predominantly leverage learning-based methodologies~\citep{li2015depth, wang2015designing, qi2018geonet, qi2020geonet++, liao2019spherical, do2020surface, bae2022irondepth, yang2024polymax}. However, direct training on real data remains challenging, as normal labels cannot be directly captured by current sensor technology. Typically, normals are thus inferred from depth maps~\citep{silberman2012indoor, eigen2015predicting}, but inaccuracies persist despite various correction strategies~\citep{bae2021estimating,long2024adaptive}. Therefore, the most recent works employed large-scale datasets of synthesised data such as OmniData~\citep{eftekhar2021omnidata, kar20223d}, along with smart architectural choices, e.g.\ per-pixel ray direction modelling in DSINE~\citep{bae2024dsine} or ViTs and diffusion models in StableNormal~\citep{ye2024stablenormal}.

\textbf{Photometric stereo}~\citep{woodham1980photometric}, on the other hand, is also a monocular technique yet it analyses multiple images captured under varying lighting. The historical inverse problem-based PS approach, which fits normals and reflectance to a physics-based image formation model, has been employed in the calibrated (known illumination) scenario~\citep{goldman2009shape,wu2011robust, shi2012biquadratic,ikehata2012robust,ikehata2014photometric}, 
in the uncalibrated one~\citep{hayakawa1994photometric,chandraker2005reflections,alldrin2007resolving,favaro2012closed,queau:2017}, 
as well as under unknown and spatially-varying lighting~\citep{basri2007photometric,queau2015tv,mo2018uncalibrated,haefner2019variational,guo2021patch}.
However, switching from the inverse problem framework to the deep learning paradigm has recently revolutionised PS. Limitations of traditional physics-based models, especially regarding non-Lambertian effects, have thereby been addressed for both calibrated  ~\citep{santo2017deep,ikehata18cnnps,chen2018ps,ikehata2022ps,ju2024deep,wei2025revisiting} and uncalibrated PS~\citep{shi2016benchmark, taniai2018neural,chen2019sdps, chen2020learned,kaya2021uncalibrated,li2022self}, and the most modern techniques make lighting context implicit, which both improves performances~\citep{ju2025revisiting} and allows tackling ``universal'' (unknown and spatially-varying illumination) PS~\citep{ikehata2022universal,ikehata23sdmunips,hardy2024unimsps, chen2025light}.

In \textbf{multi-view} single-light contexts, traditional methods, such as Colmap’s multi-view stereo \citep{schonberger2016pixelwise}, infer normal maps by analysing patch orientations \citep{bleyer2011patchmatch}. Differently, \cite{calvet2023multi} estimate normals from image grey level variations and use them in a slanted plane-sweeping algorithm. An alternative paradigm computes normals post-MVS by deriving them from depth maps, as demonstrated in RegSDF~\citep{zhang2022regsdf}, which utilises depth from Vis-MVSNet~\citep{zhang2023vis}. This depth-based normal estimation is MVS-method agnostic, applicable to both traditional~\citep{furukawa2007accurate, xu2022multi} and deep learning-driven methods~\citep{yao2018mvsnet, gu2020cascade, zhang2023vis}.

\subsection{Reflectance Estimation}

Reflectance estimation involves recovering material properties defining surface-light interactions, which is critical for realistic relighting applications. 
As for normals, various methods can be employed depending whether a single image, multiple views, or multiple illuminations are available. 

\textbf{Single-image inverse rendering} decomposes a single image into reflectance, geometry, and illumination using for instance multi-view self-supervision~\citep{yu2020self}, or diffusion models to infer physically-based rendering (PBR) materials as in  MaterialPalette~\citep{lopes2023material}. %

In \textbf{multi-view} frameworks, methods such as Ref-NeRF~\citep{verbin2022refnerf} extend neural radiance fields (NeRF) to explicitly model reflections, improving the reconstruction of specular and glossy surfaces. Techniques like SHINOBI~\citep{engelhardt23shinobi} concurrently optimise geometry, reflectance, and illumination, enabling detailed relightable 3D assets. More recently, DiffusionRenderer \citep{liang2025diffusionrenderer} provided a unified solution for high-quality normal and reflectance reconstruction from video sequences, by bridging monocular and multi-view estimation through diffusion models for neural inverse rendering.  

Finally, the aforementioned \textbf{photometric stereo} technique is the only photographic technique designed for reflectance recovery~\citep{woodham1980photometric}.
In addition, multi-light approaches offer two key advantages over single-light methods: they enhance reflectance estimation in shadowed regions (both self- and cast shadows) and improve robustness in areas affected by strong non-linearities, such as saturation in specular highlights, which shift across different lighting conditions.

\subsection{Multi-view Normal Integration}

Although individual normal maps carry valuable high-frequency geometry information, they may be inconsistent across varying viewpoints, making their fusion challenging.

In a \textbf{multi-view single-light} setting, MonoSDF~\citep{yu2022monosdf} combined a normal and depth consistency loss derived from monocular predictions \citep{eftekhar2021omnidata} with a radiance constraint, a common approach in NeRF-like \citep{mildenhall21nerf} methods.
Gaussian Surfels~\citep{dai2024high} uses monocular normal predictions to constrain 3D Gaussian Splatting (3DGS)~\citep{kerbl3Dgaussians}, while 

Yet, multi-view normal integration is much more studied under the prism of \textbf{multi-view photometric stereo}. This problem was first addressed by~\cite{EstebanVC08}, 
through the optimisation of a loss combining a rendering term and a discrepancy between photometric stereo normals and the optimised mesh. Note that this first approach required neither prior knowledge of camera poses nor illumination conditions. 
Similarly, \cite{ParkSMTK13,ParkSMTK17} %
simultaneously estimated normals, reflectance, and illumination through uncalibrated PS, leveraging the 3D mesh normals to resolve ambiguities and refining surface details. %
\cite{logothetis2019differential} later formulated the problem within an SDF representation, %
achieving superior surface detail reconstruction compared to~\citep{ParkSMTK17}.
Further refinements were made by~\cite{LiZWSDT20}, who enhanced a 3D mesh by propagating SfM points following the method of~\cite{nehab2005}, and  
validating their method through the introduction of the publicly available dataset ``DiLiGenT-MV''.

\textbf{Neural surface reconstruction methods} have recently emerged as a promising alternative, by enforcing alignment between per-view normal maps and the gradient of a neural SDF. \cite{KayaKOFG22} for instance constrained the SDF optimisation by CNN-PS normals~ \citep{ikehata18cnnps} and MVS depths~\citep{WangGVSP21}, incorporating uncertainty measures to mitigate conflicts in predictions. \cite{KayaKOFG23} further added a neural volume rendering loss, improving robustness to various material types. This resulted in a multi-objective optimisation comprising three loss terms. However, as in~\citep{KayaKOFG22}, the reliance on uncertainty-based hyperparameter tuning did not fully resolve conflicts between objectives, leading to potential loss of fine-grained details.
PS-NeRF~\citep{yang22psnerf} introduced a two-stage solution, where the first stage achieves multi-view normal integration by aligning surface gradients with SDPS-Net normals~\citep{chen2019sdps}, and the second stage leverages UNISURF~\citep{oechsle21unisurf} to optimise geometry, reflectance and illumination, modelled by multi-layer perceptrons (MLPs). However, its reliance on the directional illumination assumption limited its generalizability.
NPL-MVPS~\citep{logothetis2024nplmv} relaxed the directional illumination assumption by considering a near-light model, initially enforcing alignment between SDF gradients and UniMS-PS normals~\citep{hardy2024unimsps}. It then jointly optimises shape and reflectance via a rendering loss that explicitly models light attenuation and cast shadows. Most recently, SuperNormal~\citep{cao2024supernormal} adopted a similar initialisation approach to PS-NeRF and NPL-MVPS, utilizing SDM-UniPS normals~\citep{ikehata23sdmunips}, but significantly improved efficiency with multi-resolution hash encoding and directional finite differences, achieving nearly double the training speed.

In contrast to these prior methods, the novel approach we propose in the rest of this article formulates multi-view normal integration as a single-objective optimisation problem, through joint re-parametrisation of normals and reflectance. Let us now introduce this re-parametrisation.

\section{Normal and Reflectance Re-parametrisation}
\label{sec:reparametrisation}

Surface normals and reflectance traditionally represent distinct types of information: normals describe the geometric orientation of surfaces, while reflectance captures intrinsic photometric properties of surface materials. This inherent heterogeneity complicates their simultaneous optimisation, typically necessitating separate processing steps, e.g.,  depth triangulation from multi-view data followed by the fusion of this depth information with normal maps using multi-objective optimisation frameworks~\citep{KayaKOFG22, KayaKOFG23}.

To address this issue, we introduce a re-parametrisation approach designed to unify surface normals and reflectance into homogeneous quantities, by mapping both of them to simulated radiance values under varying illumination conditions. This radiance-based parametrisation enables a unified optimisation reducing the need for additional regularisation, commonly required in multi-objective contexts~\citep{yu2022monosdf, zhang2022regsdf}, consequently enhancing consistency and computational efficiency. Indeed, a key motivation for radiance-based re-parametrisation is its compatibility with existing photometric cost-minimisation frameworks employed in classical multi-view stereo \citep{furukawa2007accurate, schonberger2016pixelwise} and neural volume rendering methods \citep{wang21neus, yariv2021volume, li2023neuralangelo}. This incorporation facilitates optimisation of photometric consistency across viewpoints or between input and rendered images.

\subsection{Input Data}
\label{subsec:input_data}

Given a set of $N$ viewpoints, we assume the availability of corresponding normal maps $\mathsf{N}_i$ and reflectance maps $\mathsf{R}_i$, indexed by $i \in \{1, \dots, N\}$. Each viewpoint map comprises $m$ pixels, indexed by $k \in \{1, \dots, m\}$, such that: 
\begin{equation}
    \mathsf{N}_i = \{\mathbf{n}_{i,k}\}_{k \in \{1, \dots, m\}}, \quad \mathsf{R}_i = \{\mathbf{r}_{i,k}\}_{k \in \{1, \dots, m\}}.
\end{equation}
Therein, the outward normal for the $k$-th pixel in the $i$-th view is given by a unit vector $\mathbf{n}_{i,k} \in \mathbb{S}^2$, and the reflectance by a parameter vector $\mathbf{r}_{i,k} \in \mathbb{R}^q$ with dimension $q$ corresponding to the specific reflectance model used. The former is expressed, using the known camera poses, in world coordinates. If reflectance data is unavailable, simply setting $\mathbf{r}_{i,k} = [1] ~
 \forall (i,k)$ allows our framework to be used exclusively for multi-view normal integration.

\subsection{Re-parametrisation}
\label{subsec:reparametrisation}

Our re-parametrisation approach transforms each couple of normal and reflectance vectors $(\mathbf{n}_{i,k},\mathbf{r}_{i,k})$ into a vector $\mathbf{v}_{i,k}$ of homogeneous radiance values, simulated using a physically-based rendering (PBR) model:
\begin{equation} 
\mathbf{v}_{i,k} = \mathsf{F}(\mathbf{n}_{i,k}, \mathbf{r}_{i,k}, \mathsf{L}_{i,k}), 
\label{eq:reparam_general} 
\end{equation} 
with $\mathsf{L}_{i,k} \in \mathbb{R}^{n \times l}$ the illumination conditions (chosen specifically for the $k$-th photosite of the $i$-th view) and $\mathsf{F}:\, \mathbb{S}^2 \times \mathbb{R}^q \times \mathbb{R}^{n \times l} \to \mathbb{R}^{n \times q}$ the PBR function (for simplicity, the same model is used for all pixels). Here, $l$ stands for the dimensionality of the light and $n$ for that of the radiance representation, and both of them depend on the choice of a particular model.

While the general framework supports arbitrary PBR and illumination models, practical considerations regarding computational efficiency and implementation simplicity have guided our choice towards Lambertian reflectance and directional illumination. The former assumption simplifies reflectance to the albedo 
i.e., $\mathbf{r}_{i,k} \in \mathbb{R}$ or $\mathbf{r}_{i,k} \in \mathbb{R}^3$, depending on whether the images are in grey scale or RGB.
The PBR function then reads:
\begin{align}
  \mathsf{F} \colon \mathbb{S}^2 \times \mathbb{R}^q \times \mathbb{R}^{n \times 3} &\to \mathbb{R}^{n \times q} \nonumber \\
  (\mathbf{n},\mathbf{r},\mathsf{L}) &\mapsto \mathsf{F}(\mathbf{n},\mathbf{r},\mathsf{L}) = \mathsf{L} \, \mathbf{n} \, \mathbf{r}^\top,
  \label{eq:paramA} 
\end{align}
where the illumination vectors (in intensity and direction) $\mathbf{l}_{1},\dots,\mathbf{l}_{n} \in \mathbb{R}^3$ are stored row-wise in $\mathsf{L}$.

Employing $n > 3$ illumination vectors may be interesting, especially when considering more advanced PBR models (including specularity, roughness, or anisotropy) or in uncertain input scenarios, yet at the expense of bijectivity. We leave this avenue for future research.

\subsection{Optimal Illumination Directions}
\label{sec:optimal_illumination}

For this reason, we chose the widely accepted configuration suggested by \cite{Drbohlav05}, which employs three illumination configurations with equal intensity,  %
and directions spaced equally by $120^{\circ}$ in azimuth, each with a constant slant of $54.74^{\circ}$ relative to the normal vector $\mathbf{n}_{i,k}$. Alternative illumination arrangements, such as those embedded in SuperNormal~\citep{cao2024supernormal}, will be assessed in Section~\ref{subsec:supernormal_light}. %

In addition to numerical considerations, this particular choice also avoids self-shadowing, namely negative dot products between normals and illumination vectors. Nevertheless, our method could accommodate non-physical negative radiance as long as the rendering function of the downstream reconstruction technique is consistent. In the next section, we explore a first possibility for the latter downstream technique, which is surface sweeping.

\section{Surface Sweeping-based 3D Reconstruction}
\label{sec:surface_sweeping}

As a first proof of concept, let us now apply the re-parametrisation we proposed in Section~\ref{sec:reparametrisation} to triangulation-based MVS, a method that estimates the depth for each pixel in a reference view by maximizing the photometric consistency in other (control) views. In particular, we will show that by leveraging the information provided by the normals, plane sweeping-based algorithms~\citep{collins1996space} can be turned into an \textit{exact} surface sweeping-based method for integrating multi-view normal data, eliminating the inherent approximation errors due to local planar surface assumptions, e.g., fronto-parallel or slanted patches~\citep{furukawa2007accurate, bleyer2011patchmatch, schonberger2016pixelwise}.

\subsection{Objective Function}
\label{subsec:patch_objfunc}

Our aim is to compute depth values for all pixels in the first view, considered as reference. Without loss of generality, this reference camera frame aligns with the world coordinate system. The $N-1$ other views then serve as control views for enforcing multi-view consistency on the normals and reflectance. 
The optimisation of multi-view consistency is usually carried out locally, by splitting the pixels of the reference view into patches \( 
 {\mathcal{P}} \subset \mathbb{R}^2 \) containing \( m_{\mathcal{P}} \) pixels. We will denote by $\{\mathbf{v}_{1,j}\}_{j \in \{1,\dots,m_P\}}$ the re-parametrised normals and reflectance over this patch. Formulated in terms of this re-parametrisation, finding the depth of the patch centre that maximises multi-view consistency of normal and reflectance maps amounts to solving: 
\begin{equation}
    \label{eq:objective_reparam}
    \min_{z > 0} ~ \sum_{i=2}^N  \sum_{j=1}^{m_{\mathcal{P}} } \, \left\| \, \mathbf{v}_{1,j} - \mathbf{v}_{i,j}(z) \, \right\|_2^2,
\end{equation}
where  \( \mathbf{v}_{i,j}(z) \in \mathbb{R}^{3 \times q} \) stands for the re-parametrisation in the $i$-th view, sampled using bilinear interpolation at a pixel position $\mathbf{p}_{i,j}(z) \in \mathbb{R}^2$ calculated given the depth hypothesis $z$. This position is obtained by first back-projecting the reference patch pixel \( \mathbf{p}_{1,j} \in \mathcal{P} \) into 3D, and then projecting this 3D point into the control view:
\begin{equation}
    \label{eq:reprojection}
    \mathbf{p}_{i,j}(z) = \pi_i \circ \pi_{z}^{-1}(\mathbf{p}_{1,j}), \quad \forall \, \mathbf{p}_{1,j} \in \mathcal{P},
\end{equation}
where \(\pi_i\) denotes the projection from 3D world coordinates into the $i$-th control view's image plane, and \(\pi_{z}^{-1}\) the inverse projection from a reference image pixel to its corresponding 3D point, given a depth hypothesis~\(z\).

To proceed with the actual optimisation of~\eqref{eq:objective_reparam}, the function \( \pi_{z}^{-1} \) must now be made explicit, which typically involves a local geometric approximation (e.g., fronto-parallel or slanted patches). As we shall see next, the availability of normal information provides us with a natural way to avoid such an approximation.

\begin{figure*}[!ht]
  \begin{minipage}[c]{0.33\textwidth}
    \def\svgwidth{\textwidth}
    %% Creator: Inkscape 1.1.2 (0a00cf5339, 2022-02-04), www.inkscape.org
%% Accompanies image file 'FrontoPar.pdf' (pdf, eps, ps)
\begingroup%
  \makeatletter%
  \providecommand\color[2][]{%
    \errmessage{(Inkscape) Color is used for the text in Inkscape, but the package 'color.sty' is not loaded}%
    \renewcommand\color[2][]{}%
  }%
  \providecommand\transparent[1]{%
    \errmessage{(Inkscape) Transparency is used (non-zero) for the text in Inkscape, but the package 'transparent.sty' is not loaded}%
    \renewcommand\transparent[1]{}%
  }%
  \providecommand\rotatebox[2]{#2}%
  \newcommand*\fsize{\dimexpr\f@size pt\relax}%
  \newcommand*\lineheight[1]{\fontsize{\fsize}{#1\fsize}\selectfont}%
  \ifx\svgwidth\undefined%
    \setlength{\unitlength}{290.97970695bp}%
    \ifx\svgscale\undefined%
      \relax%
    \else%
      \setlength{\unitlength}{\unitlength * \real{\svgscale}}%
    \fi%
  \else%
    \setlength{\unitlength}{\svgwidth}%
  \fi%
  \global\let\svgwidth\undefined%
  \global\let\svgscale\undefined%
  \makeatother%
  \begin{picture}(1,1.00151724)%
    \lineheight{1}%
    \setlength\tabcolsep{0pt}%
    \put(0,0){\includegraphics[width=\unitlength,page=1]{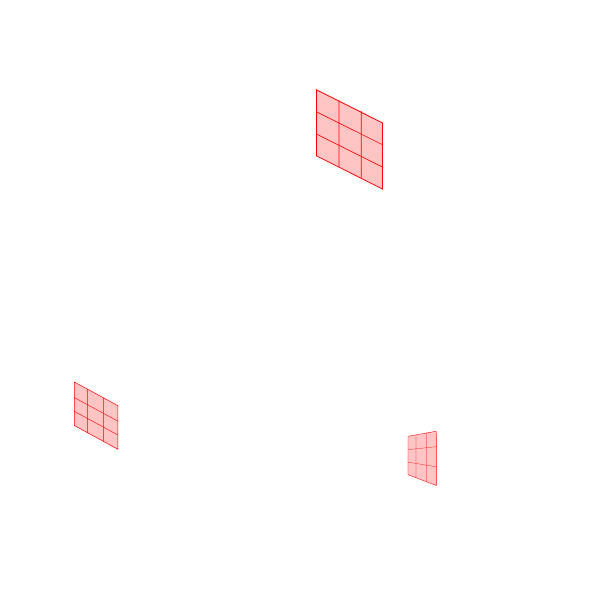}}%
    \put(0.17531435,0.26398612){\makebox(0,0)[lt]{\lineheight{1.25}\smash{\begin{tabular}[t]{l}$\mathbf{\scriptstyle{p_{1,j}}}$\end{tabular}}}}%
    \put(0.51872377,0.98417945){\makebox(0,0)[lt]{\lineheight{1.25}\smash{\begin{tabular}[t]{l}$z$\end{tabular}}}}%
    \put(0.08260399,0.36114165){\makebox(0,0)[lt]{\lineheight{1.25}\smash{\begin{tabular}[t]{l}$\small{\mathcal{P}}$\end{tabular}}}}%
    \put(-0.00138935,0.09823606){\makebox(0,0)[lt]{\lineheight{1.25}\smash{\begin{tabular}[t]{l}$\text{\footnotesize{Reference camera}}$\end{tabular}}}}%
    \put(0.54162704,0.00372161){\makebox(0,0)[lt]{\lineheight{1.25}\smash{\begin{tabular}[t]{l}$\text{\footnotesize{Control camera}}$\end{tabular}}}}%
    \put(0.72427888,0.24077035){\makebox(0,0)[lt]{\lineheight{1.25}\smash{\begin{tabular}[t]{l}$\mathbf{\scriptstyle{p_{i,j}}}$\end{tabular}}}}%
    \put(0,0){\includegraphics[width=\unitlength,page=2]{FrontoPar.pdf}}%
  \end{picture}%
\endgroup%

  \end{minipage}\hfill
  \begin{minipage}[c]{0.33\textwidth}
    \def\svgwidth{\textwidth}
    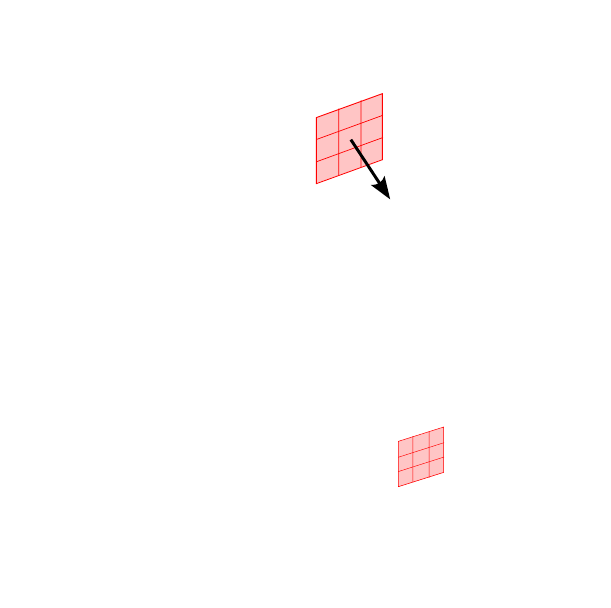
  \end{minipage}\hfill
  \begin{minipage}[c]{0.33\textwidth}
    \def\svgwidth{\textwidth}
    %% Creator: Inkscape 1.1.2 (0a00cf5339, 2022-02-04), www.inkscape.org
%% Accompanies image file 'Normal.pdf' (pdf, eps, ps)
\begingroup%
  \makeatletter%
  \providecommand\color[2][]{%
    \errmessage{(Inkscape) Color is used for the text in Inkscape, but the package 'color.sty' is not loaded}%
    \renewcommand\color[2][]{}%
  }%
  \providecommand\transparent[1]{%
    \errmessage{(Inkscape) Transparency is used (non-zero) for the text in Inkscape, but the package 'transparent.sty' is not loaded}%
    \renewcommand\transparent[1]{}%
  }%
  \providecommand\rotatebox[2]{#2}%
  \newcommand*\fsize{\dimexpr\f@size pt\relax}%
  \newcommand*\lineheight[1]{\fontsize{\fsize}{#1\fsize}\selectfont}%
  \ifx\svgwidth\undefined%
    \setlength{\unitlength}{292.47967891bp}%
    \ifx\svgscale\undefined%
      \relax%
    \else%
      \setlength{\unitlength}{\unitlength * \real{\svgscale}}%
    \fi%
  \else%
    \setlength{\unitlength}{\svgwidth}%
  \fi%
  \global\let\svgwidth\undefined%
  \global\let\svgscale\undefined%
  \makeatother%
  \begin{picture}(1,0.99638091)%
    \lineheight{1}%
    \setlength\tabcolsep{0pt}%
    \put(0,0){\includegraphics[width=\unitlength,page=1]{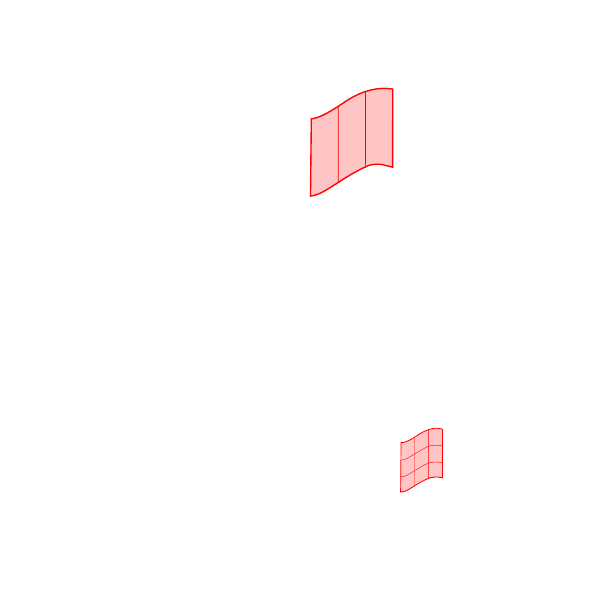}}%
    \put(0.17441525,0.26776085){\makebox(0,0)[lt]{\lineheight{1.25}\smash{\begin{tabular}[t]{l}$\mathbf{\scriptstyle{p_{1,j}}}$\end{tabular}}}}%
    \put(0.51606345,0.97913203){\makebox(0,0)[lt]{\lineheight{1.25}\smash{\begin{tabular}[t]{l}$z$\end{tabular}}}}%
    \put(0.08218034,0.35928962){\makebox(0,0)[lt]{\lineheight{1.25}\smash{\begin{tabular}[t]{l}$\small{\mathcal{P}}$\end{tabular}}}}%
    \put(-0.00138223,0.09773252){\makebox(0,0)[lt]{\lineheight{1.25}\smash{\begin{tabular}[t]{l}$\text{\footnotesize{Reference camera}}$\end{tabular}}}}%
    \put(0.53884933,0.00370252){\makebox(0,0)[lt]{\lineheight{1.25}\smash{\begin{tabular}[t]{l}$\text{\footnotesize{Control camera}}$\end{tabular}}}}%
    \put(0.72569291,0.23953557){\makebox(0,0)[lt]{\lineheight{1.25}\smash{\begin{tabular}[t]{l}$\mathbf{\scriptstyle{p_{i,j}}}$\end{tabular}}}}%
    \put(0,0){\includegraphics[width=\unitlength,page=2]{Normal.pdf}}%
  \end{picture}%
\endgroup%

  \end{minipage}\hfill
  \vspace*{1em}
  \caption{The three types of patches. From left to right: fronto-parallel patches, slanted patches and (proposed) surface patches.}
  \label{fig:patch_types}
\end{figure*}

\subsection{Exact Multi-view Integration}
\label{subsec:exact_solution}

The most common approximations used in classical plane-sweeping MVS methods, which we illustrate in the first two diagrams of Figure~\ref{fig:patch_types}, are: 
\begin{itemize}
    \item \textbf{Fronto-parallel patches} (zero-order approximation): all the patch points %
    are assumed to lie on a plane parallel to the image plane, located at the hypothesised depth \( z \). The inverse projection is then simply \( \pi_{z}^{-1}(\mathbf{p}_{1,j}) = z \, \mathsf{K}^{-1}  [\mathbf{p}_{1,j}^\top~ 1]^\top \), with $\mathsf{K}$ the intrinsics matrix of the reference camera (assumed known a priori). This is the simplest model, but it is often inaccurate for non-planar surfaces or slanted views~\citep{furukawa2007accurate,FurukawaP15}.
    \item \textbf{Slanted patches} (first-order approximation): a finer approximation retains the depth \( z \) of the patch centre, yet assumes that the other points of the patch lie on a slanted plane, whose orientation is determined by the normal of the patch centre. Their depth is thus adjusted based on the plane equation~\citep{bleyer2011patchmatch, schonberger2016pixelwise, calvet2023multi}. This generally provides better accuracy than fronto-parallel patches, but approximation errors remain for curved surfaces.
\end{itemize}

In contrast, in our framework the detailed local geometry is fully encoded in the normal map \( \mathsf{N}_{1} \), which allows us to move beyond simple planar approximations. Knowing the camera's intrinsics, the normals $\mathbf{n}_{1,j}$ in all pixels $\mathbf{p}_{1,j} \in \mathcal{P}$ can indeed 
be integrated into depth values $z$ $\alpha_j$, with $\alpha_j = 1$ in the patch center. This means that the surface patch is reconstructed up to a scale factor~\citep{queau2018normal}, which is none other than the depth $z$ of the patch centre. The inverse projection thus becomes:
\begin{equation}
\label{eq:inverse_projection_surface}
\pi_{z}^{-1}(\mathbf{p}_{1,j}) = z \, \alpha_j  \, \mathsf{K}^{-1}
    [\mathbf{p}_{1,j}^\top~ 1]^\top,
    \quad \forall \, \mathbf{p}_{1,j} \in \mathcal{P},
\end{equation}
Searching for the optimal $z$ therefore amounts to achieve ``surface sweeping'', without planar approximation.

By using this exact local surface representation, and the triangulation-based MVS framework~\eqref{eq:objective_reparam} leveraging our re-parametrisation, we not only enforce consistency in both detailed geometry (integrated normals) and appearance (reflectance) across views, but also avoid low-order approximation errors inherent to plane sweeping-based methods. This is empirically validated in the experiments presented hereafter.

\subsection{Empirical Validation}
\label{subsec:exp_res_patch}

To evaluate both the re-parametrisation loss and the surface sweeping strategy, we generated a synthetic benchmark and compared our results against those obtained using traditional methods.

\begin{figure}[htpb]
    \centering
    \begin{tabular}{cc}
         \includegraphics[width=0.45\linewidth]{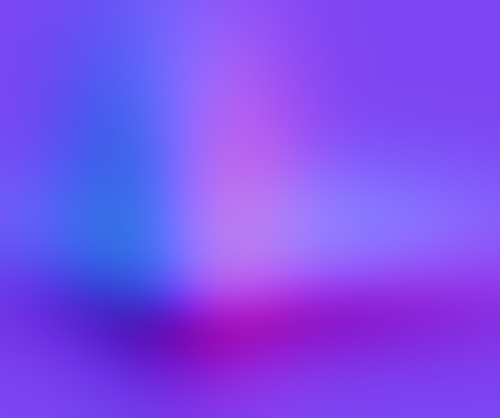} &
         \includegraphics[width=0.45\linewidth]{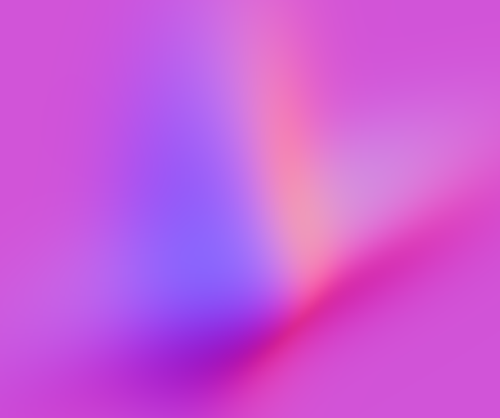}\\
         \includegraphics[width=0.45\linewidth]{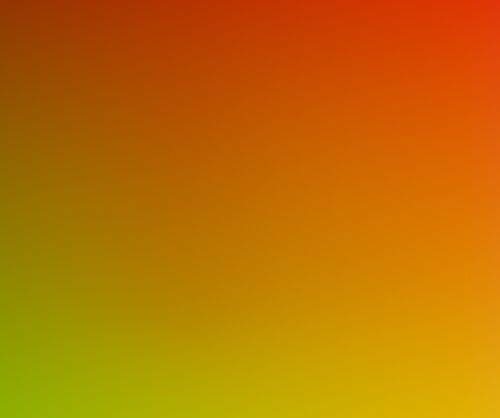} &
         \includegraphics[width=0.45\linewidth]{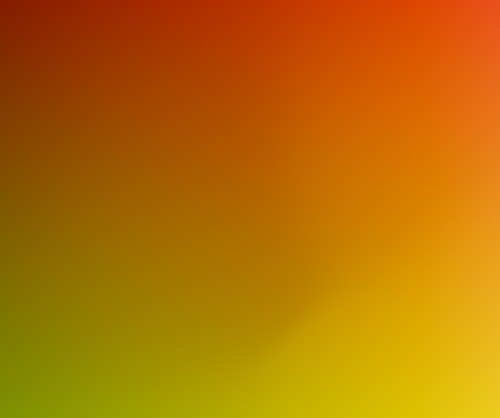}%
    \end{tabular}
    \caption{Synthetic normals (top) and reflectance (bottom) used in our experiments, for the reference view (left) and one control view (right).}
    \label{fig:synthetic_data}
\end{figure}

The \textbf{synthetic dataset} we created comprises two superimposed Gaussian functions, whose ground truth normals are computed analytically. The reflectance of the surface was set as a piecewise-linear function. %
We rendered $N = 5$ normal and reflectance maps of this surface, using the camera parameters of the first five views of the Buddha dataset from DiLiGenT-MV~\citep{LiZWSDT20}. Example normal and reflectance maps for two viewpoints are shown in Figure~\ref{fig:synthetic_data}. This dataset allows us to evaluate surface reconstruction accuracy (in terms of mean depth error over the reference view) under known, ideal conditions (no noise, perfect normal/reflectance estimates) and to study the effect of controlled noise.

\textbf{Single-objective validation} is first considered, by comparing our results against a more straightforward
solution combining a geometric loss $\mathcal{L}_{\text{geom}}$ and a reflectance discrepancy one $\mathcal{L}_{\text{photo}}$, through a weighting hyperparameter \( \mu \):
\begin{equation}
    \label{eq:objective_combined}
    \min_{z  > 0} \sum_{i=2}^N \sum_{j=1}^{m_{\mathcal{P}} } \Big(\underbrace{( 1 - \mathbf{n}_{1,j} \cdot \mathbf{ n}_{i,j}(z) )^2}_{\mathcal{L}_{\text{geom}}}  + \mu \underbrace{\| \mathbf{r}_{1,j} - \mathbf{r}_{i,j}(z) \|_2^2}_{\mathcal{L}_{\text{photo}}} \Big).
\end{equation}
Therein, tuning \( \mu \) is often difficult, since its optimal value depends on the noise level. By unifying geometric and photometric cues into a single, parameter-free objective, the proposed re-parametrisation circumvents this issue.

\begin{figure}[ht]
    \centering
    \includegraphics[width=\linewidth]{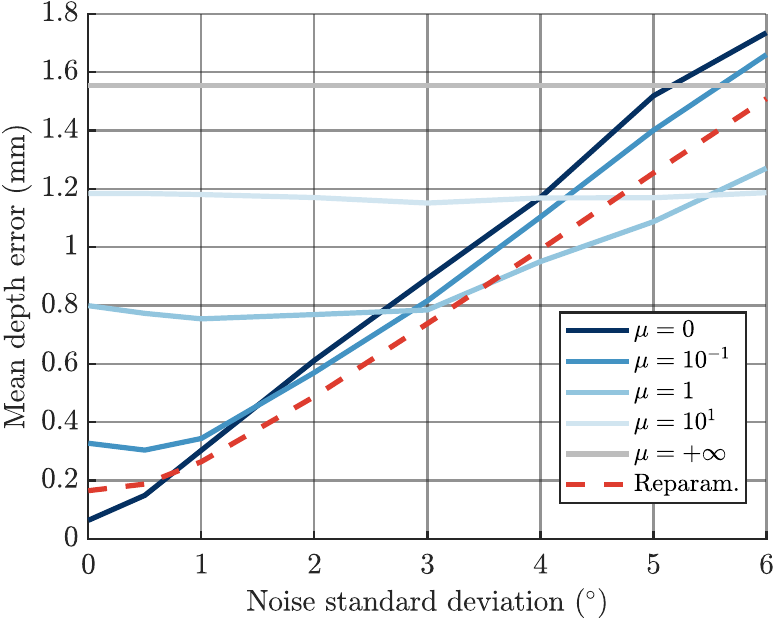}
    \caption{Mean depth estimation error %
    as a function of Gaussian noise added to input normal maps. The multi-objective approach is highly sensitive to the tuning of the hyperparameter \( \mu \), while ours maintains stable results without such tuning.}%
    \label{fig:reparam_vs_combined_loss}
\end{figure}

Figure~\ref{fig:reparam_vs_combined_loss} compares the mean depth error obtained using both approaches, in the presence of an increasing amount of additive Gaussian noise on the orientation of the normals and a fixed one on the reflectance values (standard deviation: $1\%$ of the maximum reflectance value). To focus the evaluation on the re-parametrisation, the same slanted patch approximation was adopted in both cases. While the optimal value of \( \mu \) depends on the noise level, our hyperparameter-free approach always represents a reasonable compromise in performance, regardless of the noise level. %

\begin{figure}[ht]
    \centering
    \includegraphics[width=\linewidth]{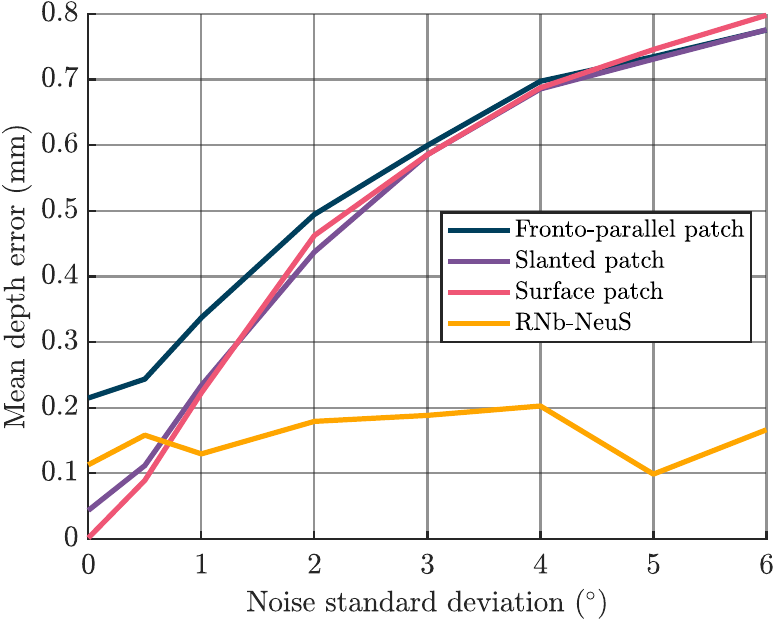}
    \caption{Mean depth estimation error as a function of Gaussian noise added to input normals, for patch-based MVS method employing fronto-parallel plane sweeping (dark blue), slanted plane sweeping (purple), and normal-aware surface sweeping (pink), as well as the volumetric rendering method RNb-NeuS~\citep{brument24rnbneus} (orange). Surface sweeping achieves exact reconstruction in the noiseless case, yet volumetric rendering is much more robust to high noise levels.}
    \label{fig:depth_error_vs_noise}
\end{figure}

To focus solely on the evaluation and comparison of sweeping methods, this time we considered only our single-objective loss, and noise was added only to normals, not to reflectance. To also compare patch-based MVS methods against more modern frameworks, we included in the comparison our previous implementation of RNb-NeuS~\citep{brument24rnbneus}, based on volumetric rendering. The results are presented in Figure~\ref{fig:depth_error_vs_noise}.

In the ideal, noiseless scenario, our proposed surface sweeping method leveraging normal integration achieves an essentially exact 3D reconstruction, demonstrating its capacity to eliminate the errors due to low-order patch geometry approximation. Such errors are indeed clearly visible on the first-order approximation  (purple curve), and even more on the zero-order one (dark blue curve). 

Adding noise to the normals reveals critical differences in robustness. The accuracy of all patch-based variants degrades as noise increases. Most notably, the surface sweeping approach, while perfect initially, rapidly deteriorates to the same level as lower-order approximations. 
This indicates that the accuracy of the local surface derived from normal integration is highly vulnerable to inaccuracies in the input normal field. %
On the other hand, volumetric rendering demonstrates considerably better robustness to high noise levels. This robustness likely benefits from its global optimisation framework and the implicit regularisation provided by the optimisation of an SDF.

Overall, these empirical findings validate our hyperparameter-free re-parametrisation strategy, and the feasibility of exact 3D reconstruction in ideal conditions. 
However, 
given that state-of-the-art photometric stereo methods often yield normals with mean angular errors in the range \([5^\circ,10^\circ]\)~\citep{hardy2024unimsps}, we believe that robustness to noise prevails over theoretical accuracy. Therefore, in the next section we turn our attention to coupling our re-parametrisation with volumetric rendering frameworks.

\section{Volume Rendering-based 3D Reconstruction}
\label{sec:volume_rendering}

In contrast with the previous section where we focused on a single depth map, the present one introduces a method for estimating a full 3D model (geometry and reflectance) that is consistent with the input normal and reflectance data. To do so, we embed the homogeneous radiance-based re-parametrisation introduced in Section~\ref{sec:reparametrisation} into a unified objective function derived from neural volume rendering (NVR) principles. The actual implementation builds upon the NeuS2 framework~\citep{wang23neus2}, yielding a highly effective method for 3D reconstruction from multi-view normal and reflectance maps.

\subsection{Surface parametrisation}
\label{subsec:surface_param_nrv}

We aim to infer a 3D model defined by two functions: a geometric map $f: \mathbb{R}^3 \to \mathbb{R}$ and a photometric map $\boldsymbol{\rho}: \mathbb{R}^3 \to \mathbb{R}^q$. Function $f$ represents the signed distance to the surface, such that surface $\mathcal{S}$ is its zero-level set: $\mathcal{S} = \{ \mathbf{x} \in \mathbb{R}^3 \,|\, f(\mathbf{x}) = 0\}$. Function $\boldsymbol{\rho}$ assigns a reflectance value to each 3D point. 
Since the actual form of $\boldsymbol{\rho}$ must be consistent with the assumptions from Section~\ref{sec:reparametrisation}, we will limit ourselves
to the Lambertian model ($q=1$, or $q=3$ for RGB data), yet extensions to more complex BRDFs are conceivable as long as they are consistent with the PBR model used for re-parametrisation.

\subsection{Objective Function}
\label{subsec:obj_func_nrv}

For clarity within this subsection, we focus on data associated with a single camera view and omit the index $i$. Input normal and reflectance values are therefore denoted simply as $\{\mathbf{n}_k\}_k$ and $\{\mathbf{r}_k\}_k$, and their joint re-parametrisation as $\{\mathbf{v}_k\}_k$.

The core idea of our NVR approach is to optimise the scene representation ($f,\boldsymbol{\rho}$) such that its rendering, under the same conditions as in Section~\ref{sec:reparametrisation}, matches the input-derived radiance vectors $\mathbf{v}_k$, by minimising:
\begin{equation}
    \mathcal{L}_{\text{NVR}}^{p}(f,\boldsymbol{\rho}) = \sum_{k=1}^m \| \tilde{\mathbf{v}}_k(f,\boldsymbol{\rho}) - \mathbf{v}_k  \|_p^p, %
\label{eq:photometric_loss_revised}
\end{equation}
with $p \in \{1,2\}$ (this choice will be discussed in the experiments), and where $\tilde{\mathbf{v}}_k(f,\boldsymbol{\rho})$ is the NVR-based radiance at pixel $k$.

The computation of the latter draws inspiration from NeuS~\citep{wang21neus}. Denoting $\mathbf{o}$ the camera centre %
and $\mathbf{d}_k$ the viewing direction associated with the $k$-th pixel, points along this ray write as $\{\mathbf{x}_k(t) = \mathbf{o} + t \, \mathbf{d}_k \, | \, t \geq 0\}$. Volume rendering then amounts to integrating individual colour contributions $\mathbf{c}$ along this ray: 
\begin{equation}\label{eq:render_init_revised} %
\tilde{\mathbf{v}}_{k}(f,\boldsymbol{\rho}) = \int^{t_{1}}_{t_{0}} \!\!\!
w(t, f(\mathbf{x}_{k}(t))) \, \mathbf{c}(\mathbf{x}_{k}(t), f, \boldsymbol{\rho}) \, \mathrm{d}t,
\end{equation}
with $[t_0, t_1]$ the integration range, and $w$ an occlusion-aware weighting function ensuring concentration around the surface~\citep{wang21neus}. 

Unlike the original NeuS which directly optimised apparent colour, %
we optimise the underlying surface properties by taking into account the PBR model $\mathsf{F}$ (it is crucial to use the same model as in~\eqref{eq:reparam_general}). Assimilating the unit outward surface normal to the gradient of the SDF, the apparent colour in~\eqref{eq:render_init_revised} writes as: 
\begin{equation}\label{eq:lnorm_final_revised} %
    \mathbf{c}(\mathbf{x}_{k}(t), f,\boldsymbol{\rho}) \!=\! \mathsf{F}\left(\nabla f(\mathbf{x}_{k}(t)),\boldsymbol{\rho}(\mathbf{x}_k(t)),\mathsf{L}_k \right),
\end{equation}
provided that $\|\nabla f\| = 1$. To simplify the optimisation process, we followed NeuS and relaxed this hard constraint into an eikonal regulariser (encouraging $\|\nabla f\|$ to be ``close'' to unity), controlled by some hyperparameter $\lambda \geq 0$:
\begin{equation}\label{eq:eikonal_revised} %
\mathcal{L}_{\text{reg}}(f) \!=\! \lambda \,\! \dfrac{\sum_{k=1}^m \! \int^{t_{\text{1}}}_{t_{0}} \! (\|\nabla f (\mathbf{x}_{k}(t)) \| \!-\! 1)^2 \, \mathrm{d}t}{m\left(t_\text{1} - t_0\right)}. %
\end{equation}

Combining \eqref{eq:render_init_revised} and~\eqref{eq:lnorm_final_revised} yields the NVR loss~\eqref{eq:photometric_loss_revised} for a single view. By averaging the contributions of all views, and adding the regularisation~\eqref{eq:eikonal_revised} as well as the same silhouette consistency regularisation as in NeuS~\citep{wang21neus},   
we obtain our complete loss function. This formulation enables end-to-end optimisation of the SDF $f$ and reflectance $\boldsymbol{\rho}$ using gradient-based methods, typically by representing $f$ and $\boldsymbol{\rho}$ as MLPs and employing hierarchical sampling along rays, similar to NeuS \citep{wang21neus}.

\subsection{SuperNormal: A Specific Case}
\label{subsec:supernormal_light}

As stated in Section~\ref{sec:reparametrisation}, in the absence of reflectance data, our framework can still be used, setting missing reflectance to constant white. If in addition, instead of the optimal triplet discussed in Section~\ref{sec:optimal_illumination}, one chooses a triplet of lights following the vectors of the canonical basis, then our framework actually comes down to a particular case which happens to be SuperNormal~\citep{cao2024supernormal}.

Indeed, using~\eqref{eq:reparam_general} and~\eqref{eq:paramA} with $\mathbf{r}_{i,k} = [1]$ and $\mathsf{L}_{i,k} = \mathbf{I}_3$, the re-parametrised inputs simplify to the input normals: $\mathbf{v}_{i,k} = \mathbf{n}_{i,k}$ $\forall (i,k)$. Similarly, using~\eqref{eq:render_init_revised} and~\eqref{eq:lnorm_final_revised} with $\boldsymbol{\rho} = \mathds{1}$ and $\mathsf{L}_k = \mathbf{I}_3$ yields the volumetric rendered normals: $\tilde{\mathbf{n}}_k(f) := \int^{t_{1}}_{t_{0}} w(t, f(\mathbf{x}_{k}(t))) \, \nabla f(\mathbf{x}_{k}(t)) \, \mathrm{d}t$. With these two simplifications and $p=2$, the NVR loss~\eqref{eq:photometric_loss_revised} becomes: 
\begin{equation}
    \mathcal{L}_\text{SN}(f) = \sum_{k=1}^m \left\| \tilde{\mathbf{n}}_k(f) - \mathbf{n}_k \right\|_2^2,
\end{equation}
which is precisely the normal consistency loss introduced in SuperNormal. Upcoming experiments, presented in Section~\ref{sec:experimental_results},  will examine the performance obtained using this particular setting, in comparison with the proposed joint optimisation over normals and reflectance using optimal triplets of light sources.

\subsection{Reflectance Singularities}
\label{subsec:reflectance_singularities}

In comparison with SuperNormal, our approach has the advantage of taking into account not only normals, but also reflectance. However, this may sometimes turn into a drawback. Indeed, the actual reflectance values weight the optimisation process, potentially leading to singularities e.g., in the presence of very dark materials. Let us demonstrate this under the Lambertian assumption, in the case of grey scale images ($q=1$). %

\begin{figure}[!ht]
\centering
\begin{tabular}{cccc}
 \includegraphics[width=0.20\linewidth]{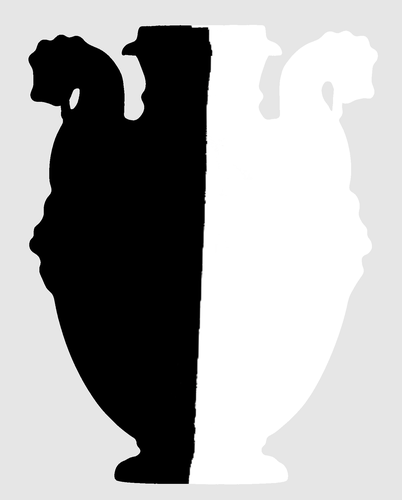} & 
 \includegraphics[width=0.20\linewidth]{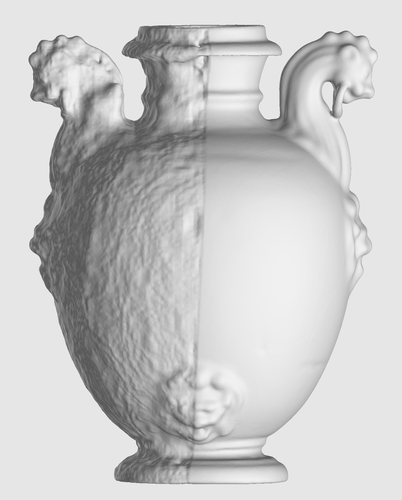} & 
 \includegraphics[width=0.20\linewidth]{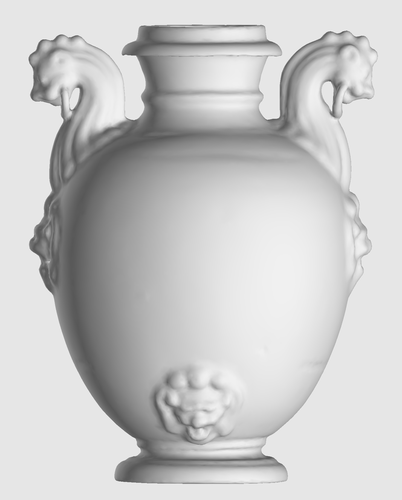} & 
 \includegraphics[width=0.20\linewidth]{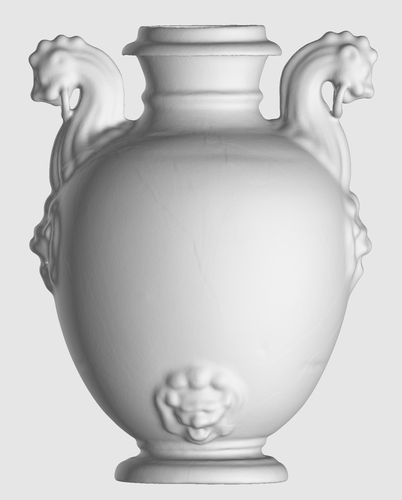} \\
(a) & (b) & (c) & (d)\\
\end{tabular}
\caption{Effect of reflectance embedding. (a) One of the input reflectance maps. (b) 3D reconstruction without reflectance embedding is deformed in the dark reflectance area. (c) With reflectance embedding, it is much closer to the ground truth (d).}
\label{fig:rgb+}
\end{figure}

For simplicity, let us consider an ideal sampling scenario where the weight $w(\cdot,f(\mathbf{x}_k(\cdot)))$ in \eqref{eq:render_init_revised} is a Dirac function $\delta_{t_k}(\cdot)$, i.e. it is null  everywhere except at some $t_k$ such that $\mathbf{x}_k:= \mathbf{x}_k(t_k)$ lies on surface $\mathcal{S}$. By combining~\eqref{eq:paramA} and~\eqref{eq:lnorm_final_revised}, the volumetric rendering~\eqref{eq:render_init_revised} then simplifies to $\tilde{\mathbf{v}}_k(f,\boldsymbol{\rho}) = \mathsf{L}_k \,  \nabla f(\mathbf{x}_k) \, \boldsymbol{\rho}(\mathbf{x}_k)^\top$.
If $q=1$, introducing the scalar variable $R$ such that $\boldsymbol{\rho}(\cdot) = \mathbf{r} \, R(\cdot)$, the NVR loss~\eqref{eq:photometric_loss_revised} can be rewritten: 
\begin{equation}
    \loss_{\text{NVR}}^p(f,R) 
    = \sum_{k=1}^{m} |\mathbf{r}_k| \left\| R(\mathbf{x}_k) \mathsf{L}_k \nabla f(\mathbf{x}_k) - \mathsf{L}_k \, \mathbf{n}_k  \right\|_p^p.
    \label{eq:rgb+}
\end{equation}
This rewriting emphasises that reflectance acts as weighting factor in the loss function. Consequently, singular reflectance values may substantially influence the optimisation of geometry. In particular, dark colours essentially leave the geometry locally unconstrained, which may result in slow, sub-optimal convergence and a loss of fine details (see Figure~\ref{fig:rgb+}-b).  

To mitigate this issue, a reflectance embedding approach is introduced, which ensures that the reflectance norm remains constant. That is, we augment the input and estimated reflectance values as $q+1$-dimensional vectors:
\begin{align}
\mathbf{r}_k^+ & := \frac{1}{q} \left[
    {\mathbf{r}_k}^\top,~ 
    \left(q-\left\|\mathbf{r}_k\right\|_p^p\right)^{1/p}\right]^\top\ , \\
\boldsymbol{\rho}^+(\cdot) & := \frac{1}{q} \left[
    \boldsymbol{\rho}(\cdot)^\top,~ 
    \left(q-\left\|\boldsymbol{\rho}(\cdot)\right\|_p^p\right)^{1/p}\right]^\top.
\end{align}
It is easy to check that the norm of these vectors %
is equal to 1, whatever the values of $p$ and $q$.
This reflectance embedding strategy %
prevents any unintended influence of singular reflectance values on geometry optimisation. This is qualitatively illustrated in Figure~\ref{fig:rgb+} on a synthetic experiment, %
using the $L_1$-norm ($p=1$) and grey level data ($q=1)$. Therein, the dark regions are not faithfully reconstructed in the standard case, whereas reflectance embedding significantly improves the results. Further experiments evaluating the interest of this embedding will be conducted in Section~\ref{sec:experimental_results}.

\subsection{From RNb-NeuS to RNb-NeuS2}
\label{subsec:rnbneus2_technical}

One of the main challenges with our initial approach, RNb-NeuS \citep{brument24rnbneus} built upon NeuS~\citep{wang21neus}, was the significant computation time required for a single reconstruction. The scene-dependent process indeed usually takes approximately 15 hours on a NVIDIA Quadro 6000. To address this issue, we restructured our algorithm to follow the NeuS2~\citep{wang23neus2} architecture. This results in a solution that is 100$\times$ faster, reducing computation time to around 5 minutes.

NeuS builds on NeRF~\citep{mildenhall21nerf} by introducing volumetric rendering with a transfer function that converts an SDF into a density function. This allows direct optimisation of scene geometry and extraction of a mesh via the SDF’s zero-level set. NeuS2 applies the same principles but replaces NeRF with Instant-NGP~\citep{mueller2022instant}, which enables real-time rendering thanks to CUDA acceleration and the use of an optimisable hash grid for sampling during volumetric rendering. While CUDA contributes the most significant performance improvement, the hash grid further reduces computation time by focusing on surface-adjacent regions, enhancing detail preservation. NeuS2 retains these key optimisations while also including the SDF-based density representation from NeuS.

To transition RNb-NeuS to the NeuS2 framework, several modifications were required. The first one involved modifying the input structure to accommodate normal and albedo maps. Unlike NeuS, which predicts colour as a function of the viewing direction, RNb-NeuS estimates albedo, which is independent of the viewing direction. Consequently, this dependency had to be removed from the inputs of the albedo prediction network. With this adjustment, albedo could be retrieved at any point $\mathbf{x}_k(t)\in \mathbb{R}^3$, allowing colour prediction for each pixel following Equation~\eqref{eq:lnorm_final_revised}. 

The next phase involved implementing the necessary loss functions. While the eikonal loss~\eqref{eq:eikonal_revised} and the silhouette consistency one remained unchanged, adapting the colour loss was more challenging. Indeed, the presence of a surface-aware PBR model in the colour prediction~\eqref{eq:lnorm_final_revised} fundamentally altered the gradient computations required for back-propagation. Unlike PyTorch, which supports automatic differentiation, NeuS2 is built entirely in CUDA, requiring manual specification of gradient computations. The details about the derivatives are in the supplementary material~\citep{suppmat}.

Now that we have outlined all the technical ingredients of the proposed method, let us present a series of experiments designed to evaluate its effectiveness across a range of scenarios.

\section{Experimental Results}
\label{sec:experimental_results}

This section presents a thorough empirical evaluation of our approach. The multi-view multi-light scenario (MVPS) is first considered. Then, we carry out an ablation study over the individual components of our framework. Eventually, we demonstrate the use of our method in a multi-view single-light (MVS) setup. Thorough qualitative and quantitative results are provided in the supplementary material~\citep{suppmat}.

\newcommand{\tableLineColor}{black}

\newcommand{\defineColor}[1]{\renewcommand{\tableLineColor}{#1}}

\newcommand{\coloredLine}[1]{\textcolor{\tableLineColor}{#1}}

\subsection{MVPS Materials}

Our approach uses normal and reflectance maps, derivable from PS techniques. Multi-view multi-light datasets are thus key, as they enable independent PS per view, providing inputs for our integration process.

\subsubsection{Evaluation Datasets}

\textbf{DiLiGenT-MV} \citep{LiZWSDT20} is a benchmark of five real-world objects, some with complex surface profiles and reflectance. %
Each object is imaged from 20 calibrated viewpoints using the classical turntable MVPS acquisition setup~\citep{EstebanVC08}, under 96 directional lighting conditions. Given the acquisition characteristics, the  relatively low image resolution of $612 \times 512$ pixels corresponds to approximately $0.4$ mm per pixel.

\textbf{LUCES-MV} \citep{logothetis2024luces} features 10 objects with a larger reflectance diversity than DiLiGenT-MV. Each object is captured from 12 different angles using a turntable setup, under 15 different near-light (non-directional) conditions. Considering the low number of views and lighting conditions, LUCES-MV can be considered a ``sparse'' dataset. %
However, the image resolution ($2080 \times 1552$ pixels) is significantly higher than that of DiLiGenT-MV. When related to the actual object size and the camera distance, this results in a finer scene resolution, where each pixel corresponds to approximately $0.13$ mm.
 
\textbf{Skoltech3D} \citep{voynov2023multi} is a multi-sensor dataset designed for multi-view surface reconstruction. It 
includes %
107 objects, among which we selected 20, focusing on the most challenging reflectance properties (transparency, specularity, uniform texture). 
The acquisitions are performed under highly challenging lighting conditions %
(LED panels significantly deviating from the directional assumption), %
and often present overexposed regions. %
For our evaluation, we used 20 views of the right camera, which is a TIS camera (\emph{the imaging source}) of industrial type, and 12 illuminations per view.  Although the image resolution is high ($2368  \times 1952$), the increased object-to-camera distance results in a scene representation of approximately $0.3$ mm per pixel, placing it between DiLiGenT-MV and LUCES-MV in terms of scene resolution.

\subsubsection{Photometric Stereo Methods}

Given the variety of reflectance properties and illumination conditions in the selected MVPS datasets,  
state-of-the-art uncalibrated PS methods were selected, %
particularly transformers-based ones which exhibit strong performance even under unknown, general lighting conditions. 

Among these methods, SDM-UniPS~\citep{ikehata23sdmunips} (referred to as SDM in the result subsections) provides both normal and reflectance maps, enabling a complete evaluation of our pipeline. UniMS-PS~\citep{hardy2024unimsps} predicts slightly more accurate normals (see Table~\ref{table:normal_mae}), yet no reflectance. It thus allows only for normal integration assessment, similar to SuperNormal~\citep{cao2024supernormal}. To establish an upper bound on the achievable reconstruction quality, experiments were also conducted using ground truth normals. 

In the forthcoming experiments, the notation N: X (resp. N/R: X) indicates that the inputs are normals alone (resp. normals and reflectance), estimated with method X. 

Since architecture constraints limit SDM-UniPS to use only 10 PS images, a sampling strategy was employed. Specifically, each $\mathbf{n}_k$ and $\mathbf{r}_k$ was computed as the median of the estimated normals and reflectances across $100$ random trials, each trial involving 10 randomly chosen images. UniMS-PS scaling better thanks to multi-scale enhancement, sampling was not necessary. 

\begin{table}[htbp]
\centering
\footnotesize
\defineColor{red}
\begin{tabular}{l c c c}
 & \multicolumn{3}{c}{\textbf{Normal MAE ($^\circ$) $\downarrow$}}\\
\toprule
& & \multicolumn{1}{c}{\textbf{SDM-UniPS}}  & \multicolumn{1}{c}{\textbf{UniMS-PS}}\\
\midrule

\textbf{DiLiGenT-MV} & \cellcolor{lightgray!30}Mean & \cellcolor{lightgray!30}6.79 & \cellcolor{lightgray!30}5.17 \\
& Std. & 6.68 & 7.01 \\
\cmidrule{1-4}
\textbf{LUCES-MV} & \cellcolor{lightgray!30}Mean & \cellcolor{lightgray!30}14.93 & \cellcolor{lightgray!30}13.24\\
 & Std. & 11.36 & 10.22 \\
\cmidrule{1-4}
\textbf{Skoltech3D} & \cellcolor{lightgray!30}Mean & \cellcolor{lightgray!30}17.25 & \cellcolor{lightgray!30}18.51 \\
& Std. & 13.95 & 14.36 \\

\bottomrule
\end{tabular}
\caption{Normal mean angular error for the two PS methods, on the three benchmarks. %
For each benchmark, `Mean' indicates the averaged error on all datasets and `Std.' its standard deviation.}
\label{table:normal_mae}
\end{table}

The assessment of PS-estimated normals, conducted in Table~\ref{table:normal_mae}, %
reveals that both methods %
exhibit a significant disparity across the datasets: while DiLiGenT-MV maintains an average MAE around~$6^{\circ}$, LUCES-MV shows a much higher error ($14^{\circ}$) and Skoltech3D an even higher one~($18^{\circ}$). %
The input normals provided to our method can therefore be considered rather noisy.

\subsubsection{Baselines}

Our approach was compared against various MVPS techniques 
including, for DiLiGenT-MV, the classical methods Park16~\citep{ParkSMTK17} and Li20~\citep{LiZWSDT20}, the neural approaches PS-NeRF~\citep{yang22psnerf}, MVPSNet~\citep{zhao23mvpsnet}, Kaya22~\citep{KayaKOFG22} and Kaya23~\citep{KayaKOFG23}, and the most recent methods SuperNormal~\citep{cao2024supernormal} and NPL-MVPS~\citep{logothetis2024nplmv}. Comparisons are more limited on LUCES-MV and Skoltech3D, as all methods were originally not benchmarked on these datasets, and are either proprietary or complex to reproduce.

Evaluations were also conducted against single-light MVS on DiLiGenT-MV. This required generating Lambertian-like images for each view, 
by computing for each viewpoint the median intensity across all lighting conditions, following the approach of~\cite{LiZWSDT20} and \cite{KayaKOFG22,KayaKOFG23}.

\subsubsection{Metrics}

\begin{figure}[!ht]
\centering
\begin{tabular}{cc}
\hspace{-3mm}
\includegraphics[height = 0.5\linewidth]{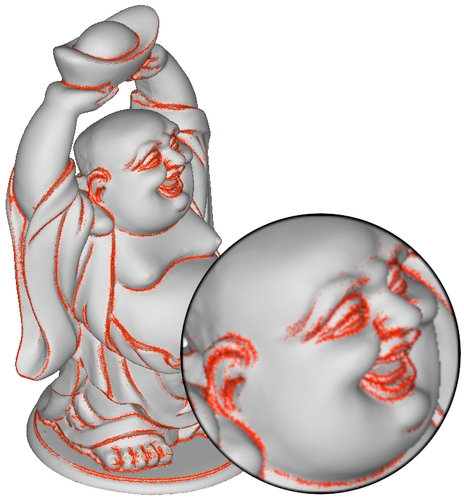} &
\includegraphics[height = 0.42\linewidth]{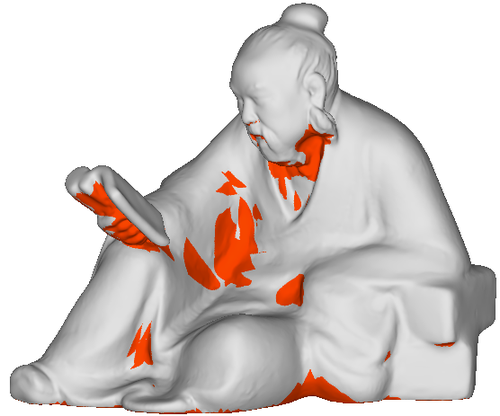} 
\end{tabular}
\caption{High curvature (left) and low visibility (right) areas, on the Buddha and Reading datasets from DiLiGenT-MV.}
\label{fig:visibility_seg}
\end{figure}

To assess both the overall reconstruction and fine details, our quantitative evaluations rely on Chamfer distance (CD) and mean angular error (MAE). %
To highlight the ability to capture fine geometric details and the robustness in poorly constrained scenarios, we also provide a focus on clusters of particular interest -- namely high curvature and low visibility areas, as illustrated in Figure~\ref{fig:visibility_seg}. 
These clusters were segmented using VCGLib~\citep{vcglib} and Meshlab~\citep{meshlab}, by retaining the vertices with maximal absolute principal curvature higher than $1.6$ and those visible in less than 5 views, respectively. 

\subsection{MVPS Results}
\label{sec:exp_mvps}

Unless otherwise specified, all subsequent evaluations are carried out using the $\mathcal{L}_{\text{NVR}}^{p}$ loss with $p=2$, an optimal lighting triplet and reflectance embedding, which has consistently shown the best performance across all evaluated datasets (see the ablation study in Section~\ref{ablation}).

\begin{table*}[htbp]
    \centering
    \footnotesize
    \begin{tabular}{l ccccc c c}
        & \multicolumn{6}{c}{\textbf{Chamfer distance (mm) $\downarrow$}} & \\
        \toprule
        \textbf{Methods} & Bear & Buddha & Cow & Pot2 & Reading & Mean & \textbf{Approx. Time} \\
        \midrule
        Park16 & 0.932 & 0.375 & 0.335 & 0.972 & 0.525 & 0.628 & N/A \\
        Li20 $\dagger$ & 0.206 & 0.257 & 0.101 & 0.213 & 0.255 & 0.206 & N/A \\
        Kaya22 & 0.380 & 0.396 & 0.297 & 0.390 & 0.344 & 0.361 & N/A \\
        PS-NeRF & 0.253 & 0.304 & 0.275 & 0.248 & 0.353 & 0.287 & $\sim$8-22h \\
        Kaya23 & 0.320 & 0.202 & 0.209 & 0.370 & 0.270 & 0.274 & N/A \\
        MVPSNet & 0.288 & 0.272 & 0.244 & 0.296 & 0.243 & 0.269 & N/A \\
        SuperNormal (N: SDM) & 0.177 & 0.206 & 0.186 & \cellcolor{bestblue!30}0.141 & \cellcolor{bestblue!30}0.222 & 0.186 & $\sim$5min \\
        NLP-MVPS & 0.211 & \cellcolor{bestgreen!30}0.176 & 0.169 & 0.213 & 0.250 & 0.204 & N/A \\
        RNb-NeuS (N/R: SDM) & 0.255 & 0.216 & 0.282 & 0.175 & 0.276 & 0.241 & $\sim$15h \\
        \midrule
        Ours (N: CNN-PS) & 0.160 & 0.197 & 0.242 & 0.166 & 0.283 & 0.210 & \multirow{6}{*}{$\sim$5min} \\
        Ours (N: SDPS-Net) & 0.182 & 0.196 & \cellcolor{bestgreen!30}0.140 & \cellcolor{bestblue!30}0.141 & 0.240 & \cellcolor{bestblue!30}0.180 & \\
        Ours (N/R: SDM) & 0.218 & 0.222 & 0.180 & 0.143 & 0.284 & 0.209 & \\
        Ours (N: SDM) & \cellcolor{bestgreen!30}0.156 & 0.219 & 0.187 & \cellcolor{bestgreen!30}0.134 & 0.276 & 0.194 & \\
        Ours (N: UniMS-PS) & \cellcolor{bestblue!30}0.157 & \cellcolor{bestblue!30}0.183 & \cellcolor{bestblue!30}0.154 & 0.148 & \cellcolor{bestgreen!30}0.191 & \cellcolor{bestgreen!30}0.167 & \\
        \rowcolor{lightgray!30}
        Ours (N: GT) & 0.127 & 0.069 & 0.097 & 0.083 & 0.090 & 0.093 & \\
        \bottomrule
    \end{tabular}
    \caption{Chamfer distance (lower is better) and approximate training times on the DiLiGenT-MV dataset. \colorbox{bestgreen!30}{Best results}. \colorbox{bestblue!30}{Second best}. Since $\dagger$ requires manual efforts~\citep{LiZWSDT20}, it is not ranked. }%
    \label{table:dlmv_cd}
\end{table*}

\begin{table*}[htbp]
\centering
\footnotesize
\begin{tabular}{l ccccc c}
& \multicolumn{6}{c}{\textbf{Normal MAE (°) $\downarrow$}}\\
\toprule
\textbf{Methods} & Bear & Buddha & Cow & Pot2 & Reading & Mean \\
\midrule
Park16 & 14.39 & 14.09 & 12.06 & 16.25 & 17.02 & 14.76 \\ 
Li20 $\dagger$ & 4.16 & 10.76 & 2.99 & 5.87 & 11.68 & 7.09 \\ 
Kaya22 & 6.44 & 14.11 & 5.87 & 9.91 & 12.42 & 9.75 \\ 
PS-NeRF & 4.82 & 10.88 & 5.62 & 6.72 & 13.75 & 8.36 \\ 
Kaya23 & 3.88 & 7.41 & 3.29 & 6.32 & 10.90 & 6.36 \\ 
MVPSNet & 5.81 & 12.04 & 6.31 & 7.47 & 12.29 & 8.78 \\ 
SuperNormal & 3.12 & \cellcolor{bestblue!30}7.14 & 3.30 & 4.28 & 10.88 & 5.75 \\ 
NLP-MVPS & 3.34 & 7.60 & \cellcolor{bestblue!30}2.57 & 4.87 & 12.56 & 6.19 \\ 
RNb-NeuS (N/R: SDM) & 3.88 & \cellcolor{bestgreen!30}7.06 & 3.73 & 4.01 & 11.11 & 5.96 \\ 
\midrule
Ours (N: CNN-PS) & 3.04 & 8.21 & 5.45 & 4.40 & 12.04 & 6.63 \\
Ours (N: SDPS-Net) & 2.85 & 7.25 & \cellcolor{bestgreen!30}2.56 & \cellcolor{bestgreen!30}3.71 & \cellcolor{bestblue!30}10.80 & \cellcolor{bestblue!30}5.44 \\
Ours (N/R: SDM) & 3.07 & 7.87 & 3.48 & 4.22 & 11.46 & 6.02 \\ 
Ours (N: SDM) & \cellcolor{bestblue!30}2.82 & 7.93 & 3.31 & \cellcolor{bestblue!30}3.94 & 11.31 & 5.86 \\ 
Ours (N: UniMS-PS) & \cellcolor{bestgreen!30}2.66 & 7.39 & 3.09 & 3.97 & \cellcolor{bestgreen!30}10.00 & \cellcolor{bestgreen!30}5.42 \\ 
\rowcolor{lightgray!30}
Ours (N: GT) & 2.10 & 2.40 & 1.89 & 1.79 & 6.86 & 3.01 \\ 
\bottomrule
\end{tabular}
\caption{Normal MAE (lower is better) averaged over all views on the DiLiGenT-MV dataset.}
\label{table:dlmv_mae}
\end{table*}

\subsubsection{Results on DiLiGenT-MV}

Table~\ref{table:dlmv_cd} presents the CD evaluation for the DiLiGenT-MV dataset. As a preliminary observation, let us remark the significant discrepancy (around $-13\%$) between our results and those from the conference paper (RNb-NeuS), using the SDM-UniPS normal and reflectance maps. This discrepancy is due to the change of norm in the NVR loss~\eqref{eq:photometric_loss_revised}, from $p=1$ to $p=2$, which severely impacts the CD scores. 
As shown in the ablation study (Table~\ref{table:ablation}), using the $\loss_{\text{NVR}}^1$ loss yields scores comparable with the conference paper.  

Another notable observation is that whatever the PS method, our results are on par with state-of-the-art, or even better. Nevertheless, combining our method with UniMS-PS yields better results ($-14\%$) than with SDM-UniPS, consistent with the higher accuracy of the former's normals (see the first row of Table~\ref{table:normal_mae}). This highlights the versatility of the proposed method, which adapts well to PS advancements.

\begin{figure*}[htpb]
    \centering
    \begin{tabular}{cc}
    \includegraphics[width = 0.42\linewidth]{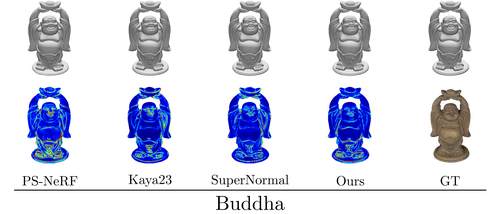} \quad & \quad
    \includegraphics[width = 0.42\linewidth]{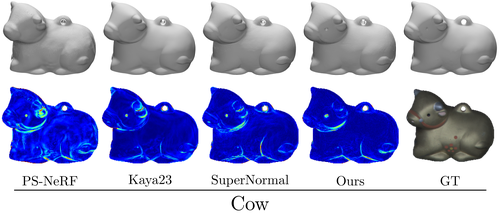} \\
    \includegraphics[width = 0.42\linewidth]{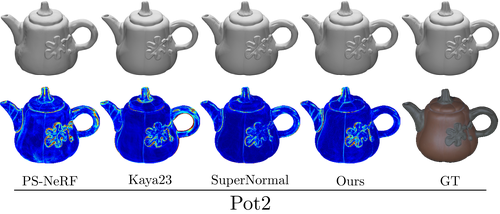} \quad & \quad
    \includegraphics[width = 0.42\linewidth]{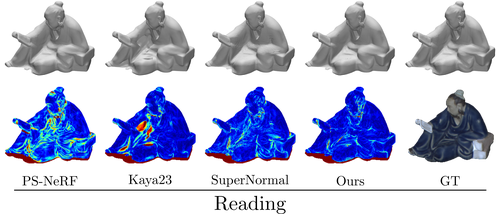} \\
    \end{tabular}
    \begin{tabular}{c}
    \includegraphics[width=0.35\textwidth]{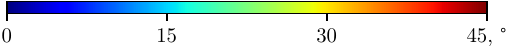}
    \end{tabular}
\caption{Reconstructed 3D mesh and corresponding MAE of four objects from DiLiGenT-MV.}
\label{fig:dmlv_mae_mesh}
\end{figure*}

The conclusions of the MAE evaluation (Table~\ref{table:dlmv_mae}) remain consistent with the CD one.
A last interesting observation is that using ground truth normals yields a non-null error, indicating a possible bias within the volumetric approach, as already noticed in Section~\ref{subsec:exp_res_patch}. 

Lastly, Figure~\ref{fig:dmlv_mae_mesh} permits qualitative assessment of the overall reconstruction and its fine details.

\subsubsection{Results on LUCES-MV}

\begin{figure*}[!ht]
    \centering
    \begin{tabular}{ccc}
        \includegraphics[height=4.6cm, keepaspectratio]{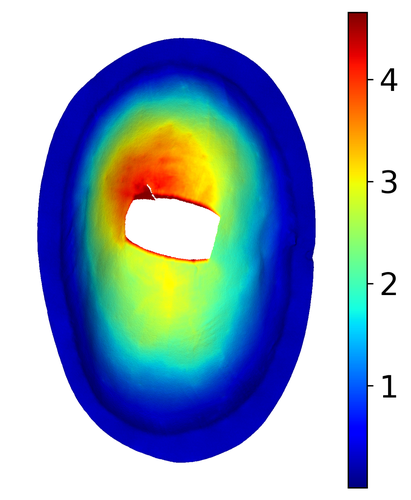} \quad & \quad
        \includegraphics[height=4.6cm, keepaspectratio]{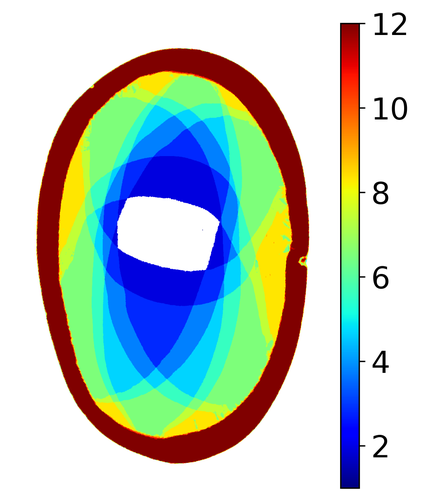} \quad & \quad
        \includegraphics[height=4.6cm, keepaspectratio]{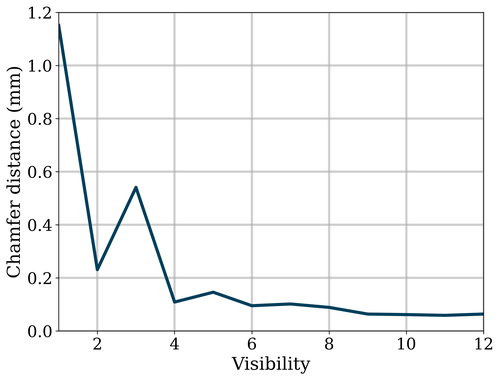}
    \end{tabular}
    \caption{Correlation between CD and visibility. From left to right: CD (in mm), number of cameras observing each vertex for LUCES-MV's Bowl object, and CD vs visibility graph over the entire dataset.}
    \label{fig:bowl_cd_vis}
\end{figure*}

Experiments on the LUCES-MV dataset (Table~\ref{table:luces_cd}) confirmed our state-of-the-art results, and here reflectance visibly improves the reconstruction. 
Interestingly, the CD scores are significantly higher in comparison with DiLiGenT-MV, despite a better pixel resolution: for the Bowl object, even using GT normals, the CD is above $0.3$ mm. We believe this could be due both to the sparser number of views — 12 vs 20 (Figure~\ref{fig:bowl_cd_vis} indeed shows a strong correlation between CD and visibility) or to inaccuracies in the dataset's camera calibration process.

\begin{table*}[htbp]
    \centering
    \footnotesize
    \begin{tabular}{lccccccccccc}
        & \multicolumn{10}{c}{\textbf{Chamfer distance (mm) $\downarrow$}}\\
        \toprule
        \textbf{Methods} & Bowl & Buddha & Bunny & Cup & Die & Hippo & House & Owl & Queen & Squirrel & Mean\\
        \midrule
        SuperNormal (N: SDM) & 0.832 & 0.828 & 0.274 & 0.770 & 0.408 & 0.421 & 0.836 & 0.586 & 0.342 & 0.301 & 0.560 \\
        \midrule
        Ours (N/R: SDM) & \cellcolor{bestblue!30}0.645 & \cellcolor{bestblue!30}0.623 & \cellcolor{bestgreen!30}0.203 & \cellcolor{bestgreen!30}0.583 & \cellcolor{bestblue!30}0.309 & \cellcolor{bestblue!30}0.309 & \cellcolor{bestblue!30}0.518 & \cellcolor{bestblue!30}0.292 & \cellcolor{bestblue!30}0.260 & 0.279 & \cellcolor{bestblue!30}0.402 \\
        Ours (N: SDM) & \cellcolor{bestgreen!30}0.624 & 0.757 & \cellcolor{bestblue!30}0.230 & \cellcolor{bestblue!30}0.609 & 0.339 & 0.335 & 0.542 & 0.321 & 0.273 & \cellcolor{bestblue!30}0.276 & 0.431 \\
        Ours (N: UniMS-PS) & 0.665 & \cellcolor{bestgreen!30}0.523 & 0.263 & 0.702 & \cellcolor{bestgreen!30}0.171 & \cellcolor{bestgreen!30}0.279 & \cellcolor{bestgreen!30}0.415 & \cellcolor{bestgreen!30}0.211 & \cellcolor{bestgreen!30}0.164 & \cellcolor{bestgreen!30}0.229 & \cellcolor{bestgreen!30}0.362 \\
        \cellcolor{lightgray!30}Ours (N: GT) & \cellcolor{lightgray!30}0.354 & \cellcolor{lightgray!30}0.152 & \cellcolor{lightgray!30}0.082 & \cellcolor{lightgray!30}0.138 & \cellcolor{lightgray!30}0.083 & \cellcolor{lightgray!30}0.099 & \cellcolor{lightgray!30}0.111 & \cellcolor{lightgray!30}0.054 & \cellcolor{lightgray!30}0.060 & \cellcolor{lightgray!30}0.073 & \cellcolor{lightgray!30}0.121 \\
        \bottomrule
    \end{tabular}
    \caption{Chamfer distance (lower is better) averaged overall all vertices on LUCES-MV.}
    \label{table:luces_cd}
\end{table*}

\subsubsection{Results on Skoltech3D}

\begin{figure}[!ht]
\centering
\begin{tabular}{cc}
\hspace{-3mm}
\includegraphics[height = 0.53\linewidth]{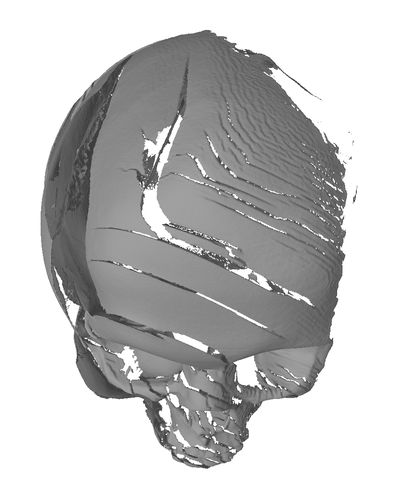} &
\includegraphics[height = 0.53\linewidth]{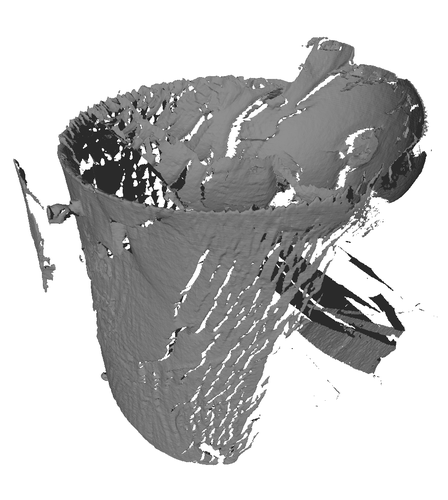} %
\end{tabular}
\caption{Reconstruction of two  objects %
from Skoltech3D. Clear surface ruptures can be observed.}
\label{fig:Skoltech3D_bad_geom}
\end{figure}

As for LUCES-MV, we compared our results on Skoltech3D against SuperNormal, which represents the state-of-the-art in MVPS. The CD reported in Table~\ref{table:skoltech_cd} emphasise the significant superiority of our results, as well as the interest of accounting for reflectance.  Nevertheless, the results on this dataset are globally disappointing, as can be seen in Figure~\ref{fig:Skoltech3D_bad_geom} on two examples exhibiting severe breaks in the surface.

\begin{table*}[!htbp]
    \centering
    \resizebox{\textwidth}{!}{
    \begin{tabular}{l cccccccccc c }
        \textbf{Methods} & 
        \rotatebox[origin=bl]{60}{dragon} & 
        \rotatebox[origin=bl]{60}{golden\_snail} & 
        \rotatebox[origin=bl]{60}{plush\_bear} & 
        \rotatebox[origin=bl]{60}{jin\_chan} & 
        \rotatebox[origin=bl]{60}{green\_carved.} & 
        \rotatebox[origin=bl]{60}{moon\_pillow} & 
        \rotatebox[origin=bl]{60}{painted\_cup} & 
        \rotatebox[origin=bl]{60}{red\_ceramic.} & 
        \rotatebox[origin=bl]{60}{painted\_sam.} & 
        \rotatebox[origin=bl]{60}{green\_tea.} &
        \rotatebox[origin=bl]{60}{blue\_boxing.}\\
       
        \toprule
        NeuS2 & 1.012 & \cellcolor{bestgreen!30}0.462 & \cellcolor{bestgreen!30}0.582 & 0.710 & 1.187 & \cellcolor{bestgreen!30}0.236 & \cellcolor{bestgreen!30}1.123 & \cellcolor{bestgreen!30}0.246 & \cellcolor{bestgreen!30}1.332 & \cellcolor{bestgreen!30}0.997 & 1.557\\
        SuperNormal (N: SDM) & 1.334 & 0.959 & 1.225 & 0.762 & 2.360 & 2.678 & 1.939 & 2.806 & 2.196 & 2.201 & 1.585 \\
        \midrule
        Ours (N/R: SDM) & \cellcolor{bestblue!30}0.984 & \cellcolor{bestblue!30}0.571 & 0.750 & \cellcolor{bestblue!30}0.491 & \cellcolor{bestgreen!30}0.545 & \cellcolor{bestblue!30}1.860 & \cellcolor{bestblue!30}1.532 & 2.049 & \cellcolor{bestblue!30}1.391 & \cellcolor{bestblue!30}1.340 & \cellcolor{bestgreen!30}1.115\\
        Ours (N: SDM) & 1.034 & 0.630 & 0.920 & \cellcolor{bestgreen!30}0.461 & \cellcolor{bestblue!30}0.602 & 2.381 & 1.710 & 2.510 & 1.714 & 1.694 & 1.288\\
        Ours (N: UniMS-PS) & \cellcolor{bestgreen!30}0.821 & 0.777 & \cellcolor{bestblue!30}0.719 & 0.530 & 0.944 & 2.084 & 1.712 & \cellcolor{bestblue!30}1.770 & 1.727 & 1.516 & \cellcolor{bestblue!30} 1.271 \\
        \cellcolor{lightgray!30}Ours (N: GT) & 
        \cellcolor{lightgray!30}0.295 & \cellcolor{lightgray!30}0.180 & \cellcolor{lightgray!30}0.171 & \cellcolor{lightgray!30}0.167 & \cellcolor{lightgray!30}0.301 & \cellcolor{lightgray!30}0.172 & \cellcolor{lightgray!30}0.471 & \cellcolor{lightgray!30}0.108 & \cellcolor{lightgray!30}0.241 & \cellcolor{lightgray!30}0.674 & \cellcolor{lightgray!30} 0.369\\

        \midrule
        & \rotatebox[origin=bl]{60}{golden\_bust} & 
        \rotatebox[origin=bl]{60}{small\_wooden.} & 
        \rotatebox[origin=bl]{60}{amber\_vase} & 
        \rotatebox[origin=bl]{60}{green\_bucket} & 
        \rotatebox[origin=bl]{60}{white\_human.} & 
        \rotatebox[origin=bl]{60}{orange\_mini.} & 
        \rotatebox[origin=bl]{60}{pink\_boot} & 
        \rotatebox[origin=bl]{60}{skate} & 
        \rotatebox[origin=bl]{60}{white\_castle.} & 
        \rotatebox[origin=bl]{60}{wooden\_trex} & 
        \rotatebox[origin=bl]{60}{Mean}\\
        \toprule
        NeuS2 &  1.173 & \cellcolor{bestgreen!30}0.300 & \cellcolor{bestgreen!30}0.769 & \cellcolor{bestgreen!30}1.693 & \cellcolor{bestgreen!30}1.630 & \cellcolor{bestgreen!30}1.641 & \cellcolor{bestgreen!30}0.639 & \cellcolor{bestgreen!30}0.908 & \cellcolor{bestgreen!30}0.760 & 1.086 & \cellcolor{bestgreen!30}0.954 \\
        
        SuperNormal (N: SDM) & 1.566 & 2.674 & 2.306 & \cellcolor{bestblue!30}1.729 & 2.179 & 2.494 & 3.580 & 3.263 & 1.302 & 1.908 & 2.050 \\
        
        \midrule
        Ours (N/R: SDM) & 0.772 & \cellcolor{bestblue!30}1.562 & \cellcolor{bestblue!30}0.872 & 2.175 & 1.667 & 2.134 & \cellcolor{bestblue!30}1.592 & \cellcolor{bestblue!30}2.450 & 1.071 & \cellcolor{bestblue!30}1.040 & \cellcolor{bestblue!30}1.332 \\
        
        Ours (N: SDM) & \cellcolor{bestgreen!30}0.686 & 2.509 & 1.599 & 2.125 & 1.774 & 2.102 & 2.980 & 2.629 & 1.157 & 1.130 & 1.602 \\
        
        Ours (N: UniMS-PS) & \cellcolor{bestblue!30}0.751 & 1.702 & 1.525 & 2.396 & \cellcolor{bestblue!30}1.655 & \cellcolor{bestblue!30}1.902 & 3.207 & 2.716 & \cellcolor{bestblue!30}0.815 & \cellcolor{bestgreen!30}0.906 & 1.497 \\
        
        \cellcolor{lightgray!30}Ours (N: GT) & \cellcolor{lightgray!30}0.218 & \cellcolor{lightgray!30}0.127 & \cellcolor{lightgray!30}0.244 & \cellcolor{lightgray!30}1.640 & \cellcolor{lightgray!30}0.269 & \cellcolor{lightgray!30}0.197 & \cellcolor{lightgray!30}0.441 & \cellcolor{lightgray!30}0.430 & \cellcolor{lightgray!30}0.289 & \cellcolor{lightgray!30}0.360 & \cellcolor{lightgray!30}0.351 \\
        
        \bottomrule
    \end{tabular}
    }
    \caption{Chamfer distance (lower is better) averaged overall all
vertices on Skoltech3D.}
\label{table:skoltech_cd}
\end{table*}

This is confirmed quantitatively by observing the strikingly high CD values. With similar spatial resolution and number of views as DiLiGenT-MV, the CD scores are drastically worse: the best-performing baseline's CD ($1.332$ mm) %
being around $6\times$ higher than for DiLiGenT-MV ($0.209$ mm). This time, we believe that it is the quality of the PS normals (in over-saturated areas, notably) that is the bottleneck -- as the very high PS errors from Table~\ref{table:normal_mae} tend to indicate.

A key indicator that this dataset is not suitable yet  %
for multi-light studies is the comparison with NeuS2, chosen as single-light MVS baseline.
NeuS2 achieves 27\% better reconstructions than MVPS-based methods using 20 views. There is thus still room for improvement in PS/MVPS research, particularly in the presence of highly challenging illumination conditions.

\subsubsection{High Curvature and Low Visibility Areas}

\begin{figure}[!ht]
    \centering
    \includegraphics[width=1.0\linewidth]{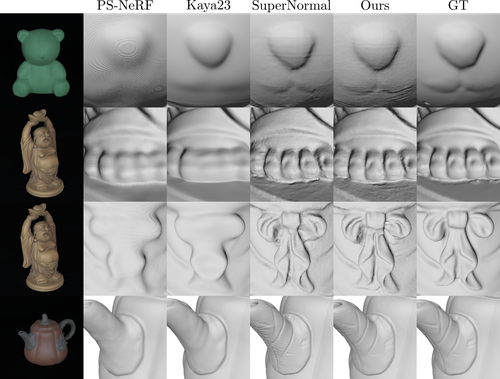}
    \caption{Focus on the high-frequency details on the DiLiGenT-MV reconstructions.}
    \label{fig:dlmv_details}
\end{figure}

As illustrated in Figure~\ref{fig:dlmv_details}, the proposed method successfully reconstructs both low- and high-frequency geometric details. However, metrics averaged over the entire  surface fail to report the accuracy in high curvature and low visibility areas. Therefore, targeted evaluations on these specific regions were conducted. The CD results on DiLiGenT-MV are reported in Table~\ref{table:table_dlmv_hclv}.

\begin{table}[!htbp]
\centering
\footnotesize
\setlength{\tabcolsep}{0.5em}
\begin{tabular}{l c c r}
& \multicolumn{3}{c}{\textbf{Chamfer distance (mm) $\downarrow$}}\\
\midrule
& \cellcolor{lightgray!20}\textbf{All} &  \cellcolor{lightgray!20}\textbf{High curv.} & \cellcolor{lightgray!20}\textbf{Diff.}\\
\midrule
\% Vertices & 100\% & 9.229\% & \\
\midrule
Park16 &  0.628 & 0.602  &  -0.026 \\
Li20 $\dagger$ &  0.206  & 0.572 &  +0.366 \\
Kaya22 &  0.361 & 0.479 &  +0.118 \\
PS-NeRF &  0.287 & 0.439 &  +0.152 \\
Kaya23 &  0.274  & 0.280  & +0.006  \\
MVPSNet & 0.269  & 0.505  &  +0.236 \\
SuperNormal (N: SDM) &  \cellcolor{bestblue!30}0.186 & 0.215 & +0.039 \\
NPL-MVPS &  0.204 & 0.258 & +0.054 \\
RNb-NeuS (N/R: SDM) & 0.241 & 0.219  & -0.022 \\
\midrule
Ours (N/R: SDM) &  0.209 & 0.218 & +0.009 \\
Ours (N: SDM) &  0.194 & \cellcolor{bestblue!30}0.210 & +0.016 \\
Ours (N: UniMS-PS) &  \cellcolor{bestgreen!30}0.167 & \cellcolor{bestgreen!30}0.206 & +0.039 \\
\cellcolor{lightgray!30}Ours (N: GT) &  \cellcolor{lightgray!30}0.093 & \cellcolor{lightgray!30}0.079 & \cellcolor{lightgray!30}-0.014 \\
\midrule
& \cellcolor{lightgray!20}\textbf{All} & \cellcolor{lightgray!20}\textbf{Low vis.} & \cellcolor{lightgray!20}\textbf{Diff.}\\
\midrule
\% Vertices & 100\% & 9.494\% & \\
\midrule
Park16 &  0.628 & 0.707  &  +0.079 \\
Li20 $\dagger$ &  0.206  & 0.894 &  +0.688 \\
Kaya22 &  0.361 & 0.610 &  +0.249 \\
PS-NeRF &  0.287 & 0.558 &  +0.271 \\
Kaya23 &  0.274  & 0.363  & +0.089  \\
MVPSNet & 0.269  & 0.571  &  +0.302 \\
SuperNormal (N: SDM) &  \cellcolor{bestblue!30}0.186 & 0.266 & +0.080 \\
NPL-MVPS &  0.204 & 0.302 & +0.098 \\
RNb-NeuS (N/R: SDM) & 0.241 & 0.286  & +0.045 \\
\midrule
Ours (N/R: SDM) &  0.209 & \cellcolor{bestblue!30}0.266 & +0.057 \\
Ours (N: SDM) &  0.194 & 0.273 & +0.079 \\
Ours (N: UniMS-PS) &  \cellcolor{bestgreen!30}0.167 & \cellcolor{bestgreen!30}0.240 & +0.073 \\
\cellcolor{lightgray!30}Ours (N: GT) &  \cellcolor{lightgray!30}0.093 & \cellcolor{lightgray!30}0.115 & \cellcolor{lightgray!30}+0.022 \\
\bottomrule
\end{tabular}
\caption{Chamfer distance on high curvature (top) and low visibility (bottom) areas, for the DiLiGenT-MV dataset.}
\label{table:table_dlmv_hclv}
\end{table}

 The key insight from this table is found in the ``Difference'' column, which compares the error for the entire point set against that of the selected subsets. As can be seen, the proposed method is significantly stabler than competitors on those challenging areas.

\subsubsection{Sparse-View Scenario}

One key advantage of using photometric stereo data is the richness of normal information. When these normals are of sufficiently high quality, as it is the case for the DiLiGenT-MV dataset, they impose strong constraints on the optimisation process, enabling the proposed approach to perform well even in the sparse-view scenario, which is highly challenging for single-light MVS methods. This is assessed in Table~\ref{table:dlmv_sparse_views}, which reports the CD evolution when reducing the number of views from 20 to 5, for a panel of recent MVS methods and for the proposed one.

\begin{table}[htbp]
\centering
\footnotesize
\setlength{\tabcolsep}{0.5em}
\begin{tabular}{l ccc}
& \multicolumn{3}{c}{\textbf{Chamfer distance (mm) $\downarrow$}}\\
\toprule
\textbf{Methods} & 20 views & 10 views & 5 views \\
\midrule
NeuS & 0.207 & 0.282 & 0.613 \\
NeuS2 & 0.331 & 0.360 & 0.529 \\
PET-NeuS & 0.820 & 0.941 & 1.069\\
GaussianSurfels & 0.973 & 1.080 & 0.894\\
RNb-NeuS (N/R: SDM) & 0.241 & 0.229 & 0.269 \\
\midrule
Ours (N/R: SDM) &  0.209 & 0.224 & 0.270 \\
Ours (N: SDM) &  \cellcolor{bestblue!30}0.194 & \cellcolor{bestblue!30}0.218 & \cellcolor{bestblue!30}0.260 \\
Ours (N: UniMS-PS) &  \cellcolor{bestgreen!30}0.167 & \cellcolor{bestgreen!30}0.183 & \cellcolor{bestgreen!30}0.217 \\
\cellcolor{lightgray!30}Ours (N: GT) &  \cellcolor{lightgray!30}0.093 & \cellcolor{lightgray!30}0.108 & \cellcolor{lightgray!30}0.118 \\
\bottomrule
\end{tabular}
\caption{Chamfer distance (lower is better) averaged overall all
vertices on DiLiGenT-MV, for a decreasing number of views.}
\label{table:dlmv_sparse_views}
\end{table}

\begin{figure*}
    \centering
    \begin{tabular}{cc}
     \includegraphics[height=0.32\linewidth]{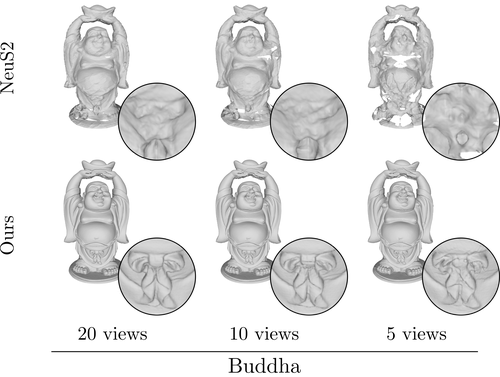}
     &
     \includegraphics[height=0.32\linewidth]{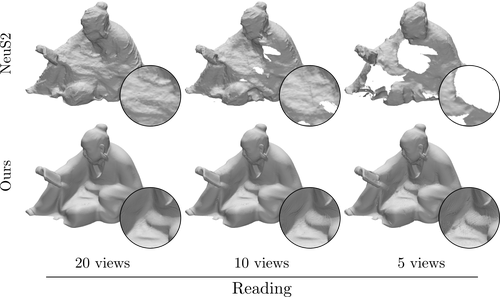}\\
    \end{tabular}
    \caption{Qualitative comparison between single-light NeuS2 and our method, in a sparse-view scenario on two objects from DiLiGenT-MV.}
    \label{fig:dlmv_sparse_view}
\end{figure*}

As expected, all MVS methods experience a significant drop in performance as the number of views decreases. %
In contrast, the proposed one remains much more stable in terms of CD. %
The primary effect of a reduced number of views for our method is a loss of fine details, as shown in Figure~\ref{fig:dlmv_sparse_view}. %

\subsection{Ablations}
\label{ablation}

Lastly, ablation studies were conducted on DiLiGenT-MV (Table~\ref{table:ablation}) and LUCES-MV (Table~\ref{table:ablation_luces}) to quantify the impact of architectural choices within our method. 
More specifically, we compared the choice of $p$ in the NVR loss $\loss_{\text{NVR}}^{p}$ (Equation~\eqref{eq:photometric_loss_revised}), of considering reflectance maps (R) in addition to normal maps, of using the pixel-wise optimal lighting triplets or the canonical ones (O vs C), and of using reflectance embedding (+) instead of a standard reflectance parametrisation.

Lighting optimisation appears important, yet to a less extent than %
the choice of $p$: %
using $\loss_{\text{NVR}}^{2}$ reduces CD scores by 9\%. %
However, %
$\loss_{\text{NVR}}^{1}$ performs better when normals and reflectance are used together, while reflectance not always contributes positively using  $\loss_{\text{NVR}}^{2}$. %
Finally, reflectance embedding (+) only slightly influences the metrics, which was to be expected since singular reflectance points are very sparse.

\begin{table}[h!]
    \setlength{\tabcolsep}{0.4em}
    \centering
    \footnotesize
    \begin{tabular}{l cccccc}
    &\multicolumn{6}{c}{\textbf{Chamfer distance (mm) $\downarrow$}} \\
    \toprule
    \textbf{Methods} & Bear & Buddha & Cow & Pot2 & Reading & Mean\\    
    \midrule
    $\loss^{2}$OR+ & 0.218 & 0.222 & 0.180 & 0.143 & 0.284 & 0.209 \\
    $\loss^{2}$CR+& 0.214 & 0.219 & 0.179 & 0.153 & 0.291 & 0.211 \\
    $\loss^{2}$OR& 0.219 & 0.230 & \cellcolor{bestblue!30}0.176 & 0.174 & 0.304 & 0.221 \\
    $\loss^{2}$CR & 0.236 & 0.221 & \cellcolor{bestgreen!30}0.173 & 0.163 & 0.312 & 0.221\\
    $\loss^{2}$O& \cellcolor{bestgreen!30}0.156 & \cellcolor{bestgreen!30}0.219 & 0.187 & \cellcolor{bestgreen!30}0.134 & \cellcolor{bestgreen!30}0.276 & \cellcolor{bestgreen!30}0.194 \\
    $\loss^{2}$C & \cellcolor{bestblue!30}0.166 & \cellcolor{bestgreen!30}0.219 & 0.181 & 0.137 & \cellcolor{bestgreen!30}0.276 & \cellcolor{bestblue!30}0.196\\
    \midrule
    $\loss^{1}$OR+& 0.212 & 0.232 & 0.255 & \cellcolor{bestblue!30}0.136 & 0.305 & 0.228 \\
    $\loss^{1}$CR+ & 0.196 & 0.235 & 0.271 & 0.164 & 0.291 & 0.231 \\
    $\loss^{1}$OR & 0.214 & 0.231 & 0.305 & 0.140 & 0.314 & 0.241 \\
    $\loss^{1}$CR & 0.185 & 0.237 & 0.301 & 0.166 & 0.315 & 0.241 \\
    $\loss^{1}$O & 0.194 & 0.234 & 0.351 & 0.160 & 0.324 & 0.252 \\
    $\loss^{1}$C & 0.183 & 0.238 & 0.388 & 0.202 & 0.326 & 0.267 \\
    \bottomrule
    \end{tabular}
    \caption{Ablation study on DiLiGenT-MV, comparing the choice of $p$ in the $\loss_{\text{NVR}}^{p}$ loss, the benefit of reflectance (R), optimal (O) or canonical (C) light triplets and finally that of reflectance embedding (+). For sake of clarity, $\loss_{\text{NVR}}^{p}$ is shown as $\loss^{p}$.}
    \label{table:ablation}
\end{table}

\begin{table*}[h!]
    \centering
    \footnotesize
    \setlength{\tabcolsep}{1em}
    \begin{tabular}{lccccccccccc}
    & \multicolumn{10}{c}{\textbf{Chamfer distance (mm) $\downarrow$}}\\
    \toprule
    \textbf{Methods} & Bowl & Buddha & Bunny & Cup & Die & Hippo & House & Owl & Queen & Squir. & Mean\\
    \midrule
    $\loss^2$OR+ & 0.645 & \cellcolor{bestblue!30}0.623 & \cellcolor{bestblue!30}0.203 & 0.583 & 0.309 & \cellcolor{bestgreen!30}0.309 & 0.518 & 0.292 & \cellcolor{bestgreen!30}0.260 & \cellcolor{bestblue!30}0.279 & \cellcolor{bestgreen!30}0.402 \\
    $\loss^2$CR+ & 0.630 & \cellcolor{bestgreen!30}0.595 & 0.226 & \cellcolor{bestgreen!30}0.563 & 0.356 & 0.376 & 0.540 & 0.312 & 0.336 & 0.365 & 0.430 \\
    $\loss^2$OR & 0.642 & 0.628 & \cellcolor{bestgreen!30}0.190 & 0.570 & 0.336 & \cellcolor{bestblue!30}0.310 & \cellcolor{bestblue!30}0.514 & \cellcolor{bestgreen!30}0.246 & 0.337 & 0.314 & \cellcolor{bestblue!30}0.409 \\
    $\loss^2$CR & 0.647 & 0.638 & 0.247 & \cellcolor{bestgreen!30}0.563 & 0.305 & 0.318 & 0.525 & \cellcolor{bestblue!30}0.269 & 0.307 & 0.387 & 0.421 \\
    $\loss^2$O& \cellcolor{bestgreen!30}0.624 & 0.757 & 0.230 & 0.609 & 0.339 & 0.335 & 0.542 & 0.321 & \cellcolor{bestblue!30}0.273 & \cellcolor{bestgreen!30}0.276 & 0.431 \\
    $\loss^2$C & \cellcolor{bestblue!30}0.626 & 0.744 & 0.230 & 0.608 & 0.340 & 0.350 & 0.566 & 0.360 & 0.296 & 0.293 & 0.441 \\
    \midrule
    $\loss^1$OR+ & 0.764 & 0.834 & 0.264 & 0.757 & \cellcolor{bestblue!30}0.295 & 0.384 & 0.524 & 0.333 & 0.308 & 0.287 & 0.475 \\
    $\loss^1$CR+ & 0.801 & 0.891 & 0.266 & 0.758 & 0.303 & 0.387 & 0.581 & 0.420 & 0.368 & 0.298 & 0.507 \\
    $\loss^1$OR & 0.762 & 0.813 & 0.235 & 0.682 & \cellcolor{bestgreen!30}0.263 & 0.356 & \cellcolor{bestgreen!30}0.489 & 0.316 & 0.295 & 0.286 & 0.450 \\
    $\loss^1$CR & 0.797 & 0.769 & 0.240 & 0.740 & 0.365 & 0.347 & 0.531 & 0.379 & 0.348 & 0.296 & 0.481 \\
    $\loss^1$O & 0.873 & 0.899 & 0.332 & 0.746 & 0.434 & 0.437 & 0.580 & 0.415 & 0.367 & 0.317 & 0.540 \\
    $\loss^1$C & 0.920 & 0.948 & 0.329 & 0.693 & 0.410 & 0.459 & 0.635 & 0.478 & 0.425 & 0.338 & 0.564 \\
    \bottomrule
    \end{tabular}
    \caption{Ablation study on LUCES-MV, with the same notations as in Table~\ref{table:ablation}.}
    \label{table:ablation_luces}
\end{table*}

\subsection{MVS Reconstruction}
\label{subsec:single_light}

In addition to the MVPS use case, we also conducted a series of experiments in the single-light scenario. Therein, no reflectance input was fed to our method, and the normals were obtained using Colmap~\citep{schonberger2016pixelwise}, which estimates them using PatchMatch~\citep{bleyer2011patchmatch}.
Additional tests were conducted using the single-view normal prediction method DSINE~\citep{bae2024dsine}, however we did not include the results due to their significantly lower quality.

\begin{figure*}[ht]
    \centering
    \begin{tabular}{ccc}
        \includegraphics[width=0.3\linewidth]{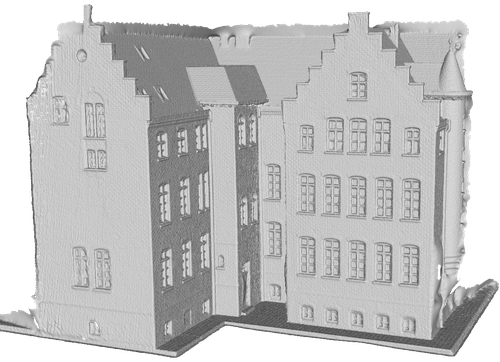} & %
        \includegraphics[width=0.3\linewidth]{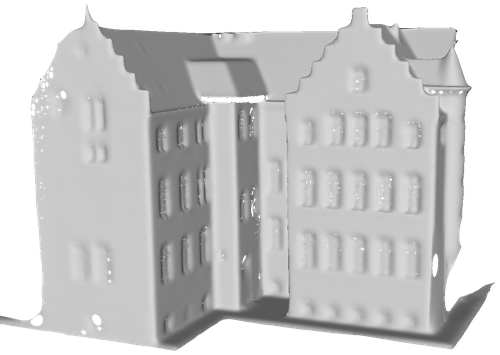} & %
        \includegraphics[width=0.3\linewidth]{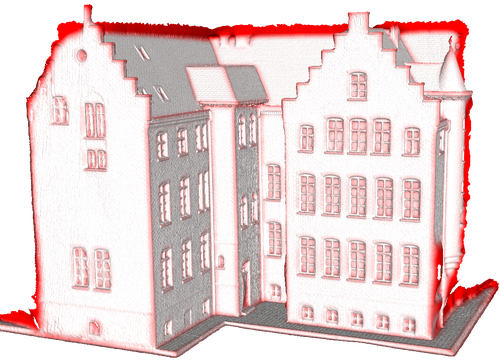} \\ %
        (a) Reference GT mesh & (b) Smoothed mesh & (c) Displacement map
    \end{tabular}
    \caption{High curvature segmentation for the scan24 object of DTU. The ground truth mesh (a), obtained via Poisson reconstruction, is smoothed using the Laplacian operator (b). The displacement magnitude (c) is then thresholded to identify the finest structures. %
    }
    \label{fig:fine_detail_eval_dtu}
\end{figure*}

\subsubsection{Methodology}

We compared our results against NeuS2~\citep{wang23neus2} and Colmap~\citep{schonberger2016pixelwise}, on 49 views of the same subset of 15 objects from DTU~\citep{jensen2014large} as in state-of-the-art methods. %

Complementing the overall Chamfer distance, we employed a specific evaluation protocol to assess the reconstruction accuracy on fine geometric details, which can be averaged out by global metrics. Therefore, as previously we focus not only on global reconstruction, but also on high curvature areas. However, the DTU objects being less ``curvy'' than other benchmarks, high curvature areas were segmented slightly differently, in order to focus on fine structures. 

Ground truth points were first meshed using screened Poisson surface reconstruction, and this mesh was then smoothed via Laplacian filtering~\citep{trimesh}. The displacement magnitude of each vertex during smoothing eventually served as a proxy for geometric frequency content, displacements above a hand-tuned threshold ($1.5$ mm) indicating fine structures. 
Figure~\ref{fig:fine_detail_eval_dtu} illustrates this segmentation approach.

\subsubsection{Results}

\begin{table*}[htbp]
\centering
\footnotesize
\begin{tabular}{l cccccccc}
& \multicolumn{8}{c}{\textbf{Chamfer distance (mm) $\downarrow$}}\\
\toprule
& \multicolumn{8}{c}{\cellcolor{lightgray!20} \textbf{All}}\\
\midrule
\textbf{Methods} & 24 & 37 & 40 & 55 & 63 & 65 & 69 & 83 \\
\midrule
Colmap &  
\cellcolor{bestblue!30}0.41 & \cellcolor{bestblue!30}0.85 &
\cellcolor{bestblue!30}0.37 & 0.37 &
\cellcolor{bestgreen!30}0.88 & 1.00 &
\cellcolor{bestgreen!30}0.52 & \cellcolor{bestgreen!30}1.17 \\
NeuS2 & 0.92 & \cellcolor{bestgreen!30}0.72 &
1.28 & \cellcolor{bestblue!30}0.30 &
0.99 & \cellcolor{bestblue!30}0.88 &
0.62 & 1.36 \\

\midrule
Ours $\loss^1$ (N:~\cite{schonberger2016pixelwise}) & 
\cellcolor{bestgreen!30}0.36 & 0.95 &
0.38 & \cellcolor{bestgreen!30}0.28 &
\cellcolor{bestblue!30}0.90 & 0.91 &
0.63 & 1.27 \\

Ours $\loss^2$ (N:~\cite{schonberger2016pixelwise}) & 
\cellcolor{bestblue!30}0.41 & 1.23 &
\cellcolor{bestgreen!30}0.36 & 0.37 &
0.95 & 1.04 &
0.73 & 1.30 \\

\midrule
& 97 & 105 & 106 & 110 & 114 & 118 & 122 & Mean\\
\cmidrule{2-9}
 & \cellcolor{bestgreen!30}1.09 & \cellcolor{bestgreen!30}0.61 &
\cellcolor{bestblue!30}0.52 & \cellcolor{bestgreen!30}0.50 &
\cellcolor{bestgreen!30}0.31 & \cellcolor{bestblue!30}0.42 &
\cellcolor{bestblue!30}0.42 & \cellcolor{bestgreen!30}0.63 \\
 & 1.62 & 0.82 &
0.58 & 0.80 &
0.37 & \cellcolor{bestgreen!30}0.40 &
\cellcolor{bestgreen!30}0.39 & 0.80 \\
\cmidrule{2-9}
& 1.33 & 0.71 &
\cellcolor{bestgreen!30}0.49 & 0.53 &
0.34 & 0.44 &
\cellcolor{bestblue!30}0.42 & \cellcolor{bestblue!30}0.66 \\
& 1.49 & 0.70 &
0.53 & 0.59 &
0.48 & 0.49 &
0.48 & 0.74 \\
\midrule
& \multicolumn{8}{c}{\cellcolor{lightgray!20} \textbf{High curvatures}}\\
\midrule
\textbf{Methods} & 24 & 37 & 40 & 55 & 63 & 65 & 69 & 83 \\
\midrule
Colmap & 
1.18 & 2.33 &
1.56 & 2.37 &
3.84 & 3.14 &
1.98 & 2.96 \\

NeuS2 & 
1.38 & \cellcolor{bestgreen!30}2.01 &
2.80 & \cellcolor{bestgreen!30}1.18 &
3.45 & 3.15 &
1.69 & 3.14 \\

\midrule
Ours $\loss^1$ (N:~\cite{schonberger2016pixelwise}) &
\cellcolor{bestgreen!30}0.88 & 2.43 & \cellcolor{bestblue!30}1.41 & \cellcolor{bestgreen!30}1.18 & \cellcolor{bestblue!30}3.10 & \cellcolor{bestgreen!30}2.50 & \cellcolor{bestgreen!30}1.61 & \cellcolor{bestgreen!30}2.86 \\

Ours $\loss^2$ (N:~\cite{schonberger2016pixelwise}) & 
\cellcolor{bestblue!30}0.96 & \cellcolor{bestblue!30}2.17 & \cellcolor{bestgreen!30}1.38 & 1.33 & \cellcolor{bestgreen!30}2.65 & \cellcolor{bestblue!30}2.62 & \cellcolor{bestblue!30}1.68 & \cellcolor{bestblue!30}2.87 \\

\midrule
 & 97 & 105 & 106 & 110 & 114 & 118 & 122 & Mean\\
\cmidrule{2-9}
 & 2.46 & 2.05 &
2.08 & \cellcolor{bestblue!30}1.88 & \cellcolor{bestgreen!30}0.91 & 1.58 & \cellcolor{bestblue!30}1.27 & 2.21 \\
& 3.31 & 2.28 &
\cellcolor{bestgreen!30}1.35 & 2.40 &
1.01 & \cellcolor{bestgreen!30}1.43 & \cellcolor{bestgreen!30}1.16 & 2.17 \\
\cmidrule{2-9}
 & \cellcolor{bestgreen!30}1.88 & \cellcolor{bestgreen!30}1.93 & \cellcolor{bestblue!30}1.52 & 1.90 & 1.08 & \cellcolor{bestblue!30}1.52 & \cellcolor{bestblue!30}1.27 & \cellcolor{bestgreen!30}1.90 \\
 & \cellcolor{bestgreen!30}1.88 & \cellcolor{bestblue!30}2.04 & 1.86 & \cellcolor{bestgreen!30}1.85 & 1.49 & 1.64 & 1.36 & \cellcolor{bestgreen!30}1.90 \\
\bottomrule
\end{tabular}
\caption{Chamfer distance on 15 objects of DTU, globally (top) and for high curvature areas only (bottom). For sake of clarity, $\loss_{\text{NVR}}^{p}$ is shown as $\loss^{p}$.}
\label{table:dtu_hc}
\end{table*}

\begin{table*}[htbp]
\centering
\footnotesize
\begin{tabular}{l ccccc}
& \multicolumn{5}{c}{\textbf{Chamfer distance (mm) $\downarrow$}}\\
\toprule
& \multicolumn{5}{c}{\cellcolor{lightgray!20}\textbf{All}}\\
\midrule
\textbf{Methods} & 49 views & 30 views & 20 views & 10 views & 5 views \\
\midrule
Colmap  & \cellcolor{bestgreen!30}0.63 & \cellcolor{bestgreen!30}0.63 & \cellcolor{bestgreen!30}0.66 & \cellcolor{bestgreen!30}1.14 & 2.11 \\
NeuS2 & 0.80 & 0.85 & 1.43  & 2.53 & 2.05 \\
\midrule
Ours $\loss^1$ (N:~\cite{schonberger2016pixelwise})  & \cellcolor{bestblue!30}0.66 & \cellcolor{bestblue!30}0.70 & \cellcolor{bestblue!30}1.15 & 1.73 & \cellcolor{bestgreen!30}1.77 \\
Ours $\loss^2$ (N:~\cite{schonberger2016pixelwise})  & 0.74 & 0.79 & 1.22 & \cellcolor{bestblue!30}1.61 & \cellcolor{bestblue!30}1.83 \\
\midrule
& \multicolumn{5}{c}{\cellcolor{lightgray!20}\textbf{High curvatures}}\\
\midrule
\textbf{Methods} & 49 views & 30 views & 20 views & 10 views & 5 views \\
\midrule
Colmap  & 2.21 & 2.37 & 2.66 & 3.35 & 4.49 \\
NeuS2 & 2.17 & 2.26 & 2.45  & 2.93 & 4.03 \\
\midrule
Ours $\loss^1$ (N:~\cite{schonberger2016pixelwise})  & \cellcolor{bestgreen!30}1.90 & \cellcolor{bestblue!30}1.97 & \cellcolor{bestblue!30}2.10 & \cellcolor{bestblue!30}2.72 & \cellcolor{bestblue!30}3.70 \\
Ours $\loss^2$ (N:~\cite{schonberger2016pixelwise})  & \cellcolor{bestblue!30}1.90 & \cellcolor{bestgreen!30}1.94 & \cellcolor{bestgreen!30}2.10 & \cellcolor{bestgreen!30}2.72 & \cellcolor{bestgreen!30}3.47 \\ 
\bottomrule
\end{tabular}
\caption{Chamfer distance on DTU averaged overall all
vertices (top) and high curvature areas (bottom), for a decreasing number of views. For sake of clarity, $\loss_{\text{NVR}}^{p}$ is shown as $\loss^{p}$.}
\label{table:dtu_cd_sparse_views}
\end{table*}

The CD values on DTU are reported in Table~\ref{table:dtu_hc}. As can be observed, the proposed method is globally on par with alternative MVS frameworks. 

However, our results are much stabler in the sparse-view scenario, as indicated in Table~\ref{table:dtu_cd_sparse_views}, where the number of views is decreased from 49 to 5.  Our results are also much more accurate in high curvature areas, as indicated in these tables. Notably, although we use Colmap's normals, our CD are $10\%$ lower in these areas, which highlights the interest of the proposed normal integration framework. Overall, these MVS experiments provide a proof of concept for the incorporation of our framework beyond MVPS. %

\section{Conclusion and Perspectives}
\label{sec:conclusion}

We presented a versatile and efficient framework for high-fidelity 3D surface reconstruction by leveraging multi-view normal cues, with optional use of reflectance information. Through a novel re-parametrisation of normals and reflectance as radiance vectors simulated under varying illumination, our method fits seamlessly into both traditional MVS and NVR pipelines. It is also very fast, thanks to CUDA acceleration and an optimisable hash grid. %
Extensive experiments on the most recent benchmarks %
demonstrate state-of-the-art performance, particularly in capturing fine details and handling challenging reflectance and visibility conditions.

However, challenges remain. We observe bias in the volume rendering version when using ground truth normals, and robustness to noisy input normals %
still needs improvement. 

Moreover, our findings suggest the need for more reliable MVPS datasets. Notably, reconstruction accuracy paradoxically decreases with increasing image resolution on current benchmarks (see for example results on DiLiGenT-MV and LUCES-MV), an observation that questions the ground truth quality of recent datasets. 

\begin{figure}[h]
    \centering
    \begin{tabular}{cc}
        \includegraphics[height=1.9cm]{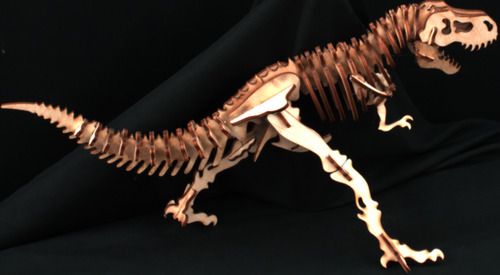} &
        \includegraphics[height=1.9cm]{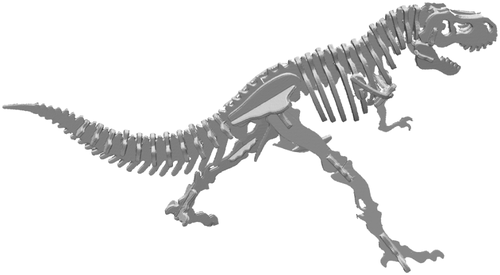} \\
        \small (a) Input image & \small (b) Reference GT mesh \\
        \includegraphics[height=1.9cm]{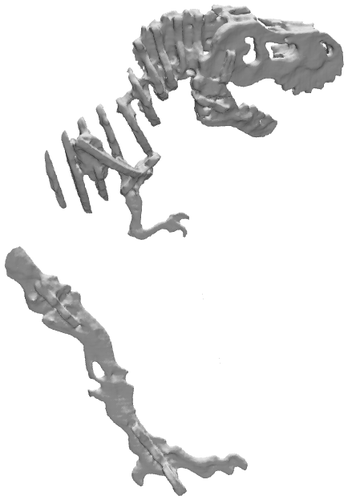} &
        \includegraphics[height=1.9cm]{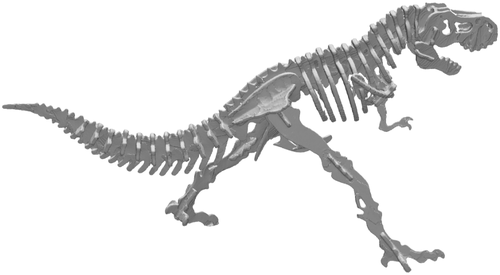} \\
        \small (c) NeuS2 & \small (d) Ours \\
    \end{tabular}
    \caption{Advantage of MVPS (d) over MVS (c) beyond fine detail reconstruction: handling deep concavities (\texttt{wooden\_trex} from the Skoltech3D dataset) associated to low visibility areas.}
    \label{fig:trex_case}
\end{figure}

\begin{figure}[h]
    \centering
    \begin{tabular}{cc}
        \includegraphics[height=1.9cm]{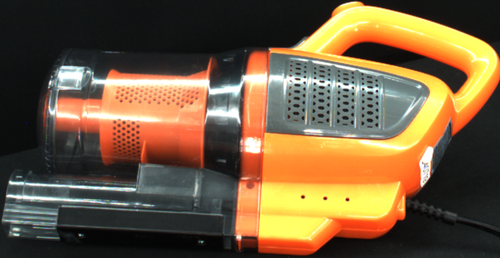} &
        \includegraphics[height=1.9cm]{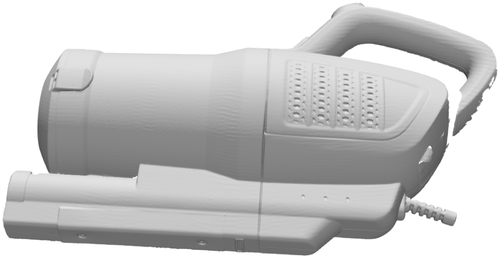} \\
        \small (a) Input image & \small (b) Reference GT mesh \\
        \includegraphics[height=1.9cm]{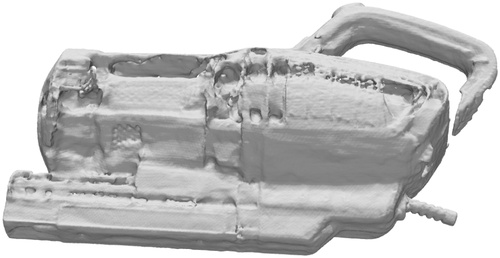} &
        \includegraphics[height=1.9cm]{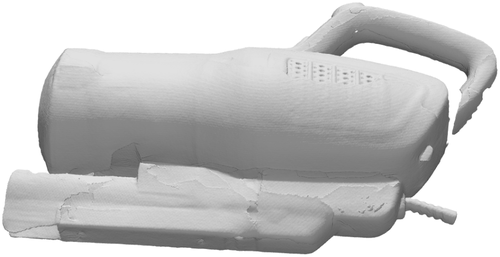} \\
        \small (c) NeuS2 & \small (d) Ours \\
    \end{tabular}
    \caption{Advantage of MVPS (d) over MVS (c) beyond fine detail reconstruction: handling transparency (\texttt{orange\_mini\_vacuum} from the Skoltech3D dataset; see transparent cover). This is enabled by the high diversity of light reflections, which are effectively leveraged by recent PS techniques.}
    \label{fig:vacum_case}
\end{figure}

More broadly, this work highlights the practical promise of MVPS over traditional MVS and single-light methods. With recent PS techniques (e.g., SDM-UniPS, UniMS-PS) 
delivering accurate normals under complex reflectance, and with multi-view normal integration methods (e.g., SuperNormal, ours) preserving fine details, MVPS emerges as a compelling option for accurate, affordable 3D scanning -- especially in semi-controlled environments. It also significantly outperforms single-illumination methods on objects with deep concavities (see Figure \ref{fig:trex_case}) or transparent parts (see Figure \ref{fig:vacum_case}), by leveraging the rich geometric cues offered by multi-light and ViT-based priors.

We hope this work encourages further exploration of this paradigm, from improving robustness and dataset quality to enabling broader applicability in real-world scenarios.

\clearpage
\subsection*{Funding}
Robin Bruneau's postdoctoral fellowship was funded by the Department of Quantitative Biomedicine at the University of Zurich.
Baptiste Brument's doctoral student fellowship was funded by the CNRS through the OPEN-DOPAMIn project.
Lilian Calvet's postdoctoral fellowship was supported by the University of Zurich, the University Hospital of Balgrist, and the OR-X - a Swiss national research infrastructure for translational surgery.
The research leading to these results received funding from the French National Research Agency (ANR) through the LabCom project ALICIA-Vision, and by the Department of Computer Science at the University of Copenhagen (DIKU) through the Copenhagen Data+ project PHYLORAMA.

\subsection*{Data Availability}
The code and data relative to this article are available at \url{https://github.com/RobinBruneau/RNb-NeuS2}.
Also, a thorough qualitative and quantitive study is available in the supplementary material \citep{suppmat} at \url{https://drive.google.com/file/d/1KDfCKediXNP5Os954TL_QldaUWS0nKcD/view?usp=drive_link}.

The DiLiGenT-MV dataset \citep{LiZWSDT20} is available at \url{https://sites.google.com/site/photometricstereodata/mv}.
The LUCES-MV dataset \citep{logothetis2024luces} is available at \url{https://drive.google.com/drive/folders/1634yweYUpLvNPC1qEG8hRpmhtxVtFrLi?usp=sharing}.
The Skoltech3D dataset \citep{voynov2023multi} is available at \url{https://skoltech3d.appliedai.tech/}.

\clearpage
\bibliography{cvpr_biblio.bib}%

\end{document}